\documentclass[10pt,twocolumn,letterpaper]{article}

\usepackage[pagenumbers]{cvpr} 

\usepackage{amsmath}
\usepackage{amssymb}
\usepackage{multicol}
\usepackage{multirow}
\usepackage{booktabs}
\usepackage{enumitem}
\usepackage[misc]{ifsym}

\definecolor{cvprblue}{rgb}{0.21,0.49,0.74}
\usepackage[pagebackref,breaklinks,colorlinks,allcolors=cvprblue]{hyperref}

\def\paperID{12484} 
\def\confName{CVPR}
\def\confYear{2025}

\title{GlyphMastero: A Glyph Encoder for High-Fidelity Scene Text Editing}

\author{
   Tong Wang\textsuperscript{\rm 1,\dag},
   Ting Liu\textsuperscript{\rm 1,\dag},
   Xiaochao Qu\textsuperscript{\rm 1},
   Chengjing Wu\textsuperscript{\rm 1},
   Luoqi Liu\textsuperscript{\rm 1,\Letter},
   Xiaolin Hu\textsuperscript{\rm 2,\Letter}
\\ 
\small \textsuperscript{\rm 1} MT Lab, Meitu Inc., Beijing 100083, China \\
\small \textsuperscript{\rm 2} Department of Computer Science and \small Technology, \\
\small BNRist, IDG/McGovern Institute for Brain Research, \\
\small Tsinghua University, Beijing 100084, China \\
\small \texttt{\{wt6, lt, qxc, ethan, llq5\}@meitu.com} \\
\small \texttt{xlhu@tsinghua.edu.cn} \\
\vspace{1mm}
\small \textsuperscript{$\dagger$}Joint first authors. \small \hspace{3mm} \textsuperscript{\Letter\ }Joint corresponding authors.
\vspace{-0.69cm} 
}

\begin{document}
\maketitle

\begin{abstract}

Scene text editing, a subfield of image editing, requires modifying texts in images while preserving style consistency and visual coherence with the surrounding environment. While diffusion-based methods have shown promise in text generation, they still struggle to produce high-quality results. These methods often generate distorted or unrecognizable characters, particularly when dealing with complex characters like Chinese. In such systems, characters are composed of intricate stroke patterns and spatial relationships that must be precisely maintained. We present GlyphMastero, a specialized glyph encoder designed to guide the latent diffusion model for generating texts with stroke-level precision. Our key insight is that existing methods, despite using pretrained OCR models for feature extraction, fail to capture the hierarchical nature of text structures - from individual strokes to stroke-level interactions to overall character-level structure. To address this, our glyph encoder explicitly models and captures the cross-level interactions between local-level individual characters and global-level text lines through our novel glyph attention module. Meanwhile, our model implements a feature pyramid network to fuse the multi-scale OCR backbone features at the global-level. Through these cross-level and multi-scale fusions, we obtain more detailed glyph-aware guidance, enabling precise control over the scene text generation process. Our method achieves an 18.02\% improvement in sentence accuracy over the state-of-the-art multi-lingual scene text editing baseline, while simultaneously reducing the text-region Fréchet inception distance by 53.28\%. 

\end{abstract}
\section{Introduction}
\label{sec:intro}

\begin{figure}[h]
\centering
\includegraphics[width=1.0\columnwidth]{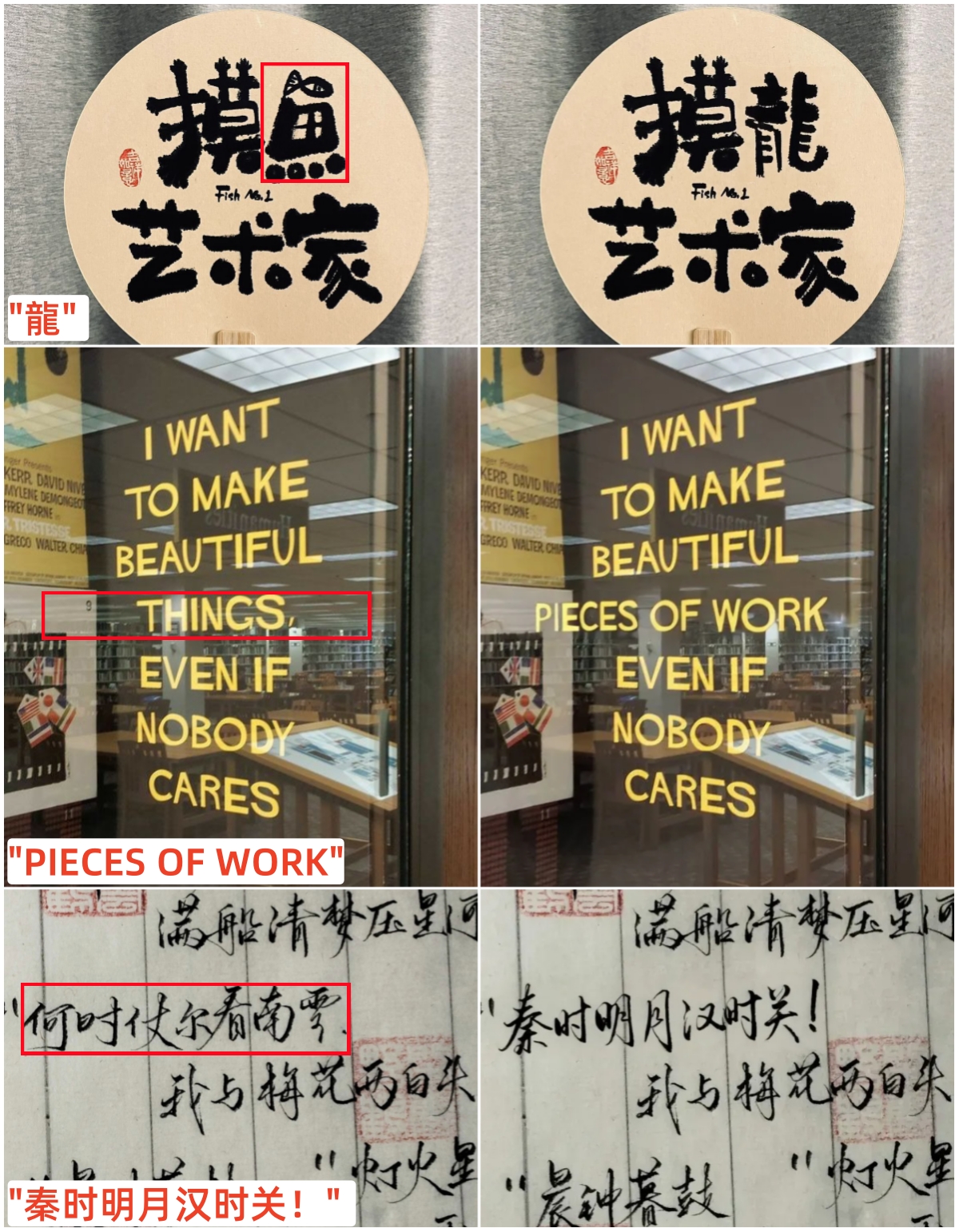}
\caption{Example results of our scene text editing method on random images collected from the internet. The original images (left column) show source text regions marked with red boxes, with target texts displayed at the bottom left. The edited results (right column) demonstrate our method's ability to preserve both structural accuracy and style consistency through stroke-level guidance across different visual styles and writing systems.}
\label{fig1}
\end{figure}

Text editing is a task that, given a selected text region, replaces existing text content with new user inputs. The most straightforward approach involves a two-stage process: text removal followed by font-matched text insertion. While this approach may suffice for images with printed texts, it requires substantial expertise in font recognition to maintain style consistency. Moreover, for text in natural scenes (i.e. scene texts), this approach faces great challenges: even with perfect font identification, achieving visual coherence between the inserted text and the scene remains infeasible due to complex environmental factors such as perspective distortions, lighting conditions, and surface properties.

This challenge motivates the task of scene text editing, a special task within image editing domain that aims to modify texts in images while preserving the original style and maintaining visual coherence with the surrounding environment. Advances in this field benefits both everyday users and professional designers, particularly when they deal with typefaces that are either difficult to identify or impossible to replicate through conventional means.

Toward this end, diffusion-based~\cite{DBLP:conf/nips/HoJA20} scene text editing methodologies~\cite{DBLP:conf/nips/ChenHL0CW23, DBLP:journals/corr/abs-2304-05568, DBLP:conf/nips/ChenXGLZLMZW23, DBLP:conf/iclr/TuoXHGX24} have demonstrated significant potential to meet these demanding requirements. DiffUTE~\cite{DBLP:conf/nips/ChenXGLZLMZW23}, a specialized scene text editing framework built upon diffusion models, exemplifies this approach. In contrast to text-to-image generation approaches that utilize text encoders such as CLIP~\cite{DBLP:conf/icml/RadfordKHRGASAM21} for extracting text embeddings from descriptive languages to enable language-driven generation, the framework employs features extracted by a pretrained OCR model from rendered text lines as the glyph guidance for scene text editing. While DiffUTE achieves robust style preservation through its conditional inpainting formulation and demonstrates effective text editing capabilities for scenes with simple printed text, it shows significant limitations in text legibility when handling complex glyph structures such as Chinese characters (Figure~\ref{fig:method_comparison}). These limitations stem from the insufficient representational capacity of its OCR feature utilization in encoding intricate glyph structures.

To address the challenge of creating robust glyph representations for achieving precise scene text editing results,  we propose \emph{GlyphMastero}, a novel trainable glyph encoder that generates fine-grained glyph guidance for scene text editing. Our glyph encoder distinguishes itself through its explicit modeling of hierarchical relationships between local character-level glyph features and global text-line structures. By capturing these hierarchical relationships and using them for glyph feature generation, our approach enables more nuanced glyph representation learning. Additionally, we incorporate multi-scale OCR features through a feature pyramid network (FPN)~\cite{DBLP:conf/cvpr/LinDGHHB17} to enhance the representation of global text structures.

We adopt the inpainting-based formulation similar to DiffUTE as our generation backbone and train it with the glyph guidance from our \emph{GlyphMastero}, achieving substantial performance improvements across writing systems ranging from Latin scripts to complex logographic systems such as Chinese characters. Both quantitative and qualitative evaluations validate the effectiveness of our approach. In Figure~\ref{fig1}, we present editing results on various images collected from the internet.
\section{Related Work}
Existing approaches for scene text editing can be broadly categorized into two paradigms: \emph{GAN-based methods} and \emph{diffusion-based methods}, each employing different strategies for conditioning the generative models.

\subsection{GAN-Based Methods}

Early approaches in the domain of scene text editing were predominantly based on Generative Adversarial Networks (GANs)~\cite{DBLP:conf/nips/GoodfellowPMXWOCB14}. Among these early works, STEFANN~\cite{DBLP:conf/cvpr/RoyBG020} introduced a dual-model architecture that combined a font-adaptive model for maintaining structural integrity with a dedicated style transfer network. In a different approach, MOSTEL~\cite{DBLP:conf/aaai/QuTXXW023} advanced the field by incorporating explicit stroke guidance maps to precisely delineate editing regions, while implementing a semi-supervised learning strategy that leveraged both synthetic and real-world data to enhance the model's generalization capabilities. In parallel, several other GAN-based methodologies~\cite{DBLP:journals/corr/abs-2107-11041, DBLP:conf/mm/WuZLHLDB19, DBLP:journals/corr/abs-2207-09649, DBLP:conf/cvpr/YangHL20} explored various architectural modifications to improve the quality of text generation. Nevertheless, these GAN-based approaches consistently encountered significant challenges when dealing with complex text structures and diverse style variations, frequently resulting in generated text that exhibited unrealistic characteristics and compromised visual quality.

\subsection{Diffusion-Based Methods}

Diffusion models have recently shown superior capabilities in modeling complex data distributions, making them well-suited for scene text editing tasks. These methods primarily differ in their conditioning strategies, which can be categorized into two types: cross-attention guidance and latent space guidance.

Cross-attention mechanisms have proven effective for conditioning diffusion models with additional information in scene text editing. DiffSTE~\cite{DBLP:journals/corr/abs-2304-05568} employs a character encoder for glyph information alongside CLIP~\cite{DBLP:conf/icml/RadfordKHRGASAM21} encoded instructions. AnyText~\cite{DBLP:journals/corr/abs-2311-03054} integrates OCR-extracted neck features from glyph images with CLIP-encoded descriptions. DiffUTE~\cite{DBLP:conf/iclr/TuoXHGX24} utilizes fixed-length features from TrOCR's~\cite{DBLP:conf/aaai/LiLC0LFZ0W23} final hidden state. Building on DiffUTE, we replace text embeddings with glyph guidance through our novel trainable glyph encoder for comprehensive glyph representations.

Latent space guidance, popularized by ControlNet~\cite{DBLP:conf/iccv/ZhangRA23}, introduces conditioning signals into diffusion models by encoding input conditions through convolutional layers and incorporating them into latent representations.~Several works in text generation~\cite{DBLP:conf/nips/ChenHL0CW23, DBLP:journals/corr/abs-2303-17870, DBLP:conf/nips/YangGYLDH023, DBLP:journals/corr/abs-2311-03054} have explored adapting this approach for scene text editing. TextDiffuser~\cite{DBLP:conf/nips/ChenHL0CW23} implements character-level segmentation masks for fine-grained control, while GlyphControl~\cite{DBLP:conf/nips/YangGYLDH023} and AnyText~\cite{DBLP:journals/corr/abs-2311-03054} introduce indirect latent conditioning through feature concatenation. While these methods demonstrate success in text generation, their reported results suggest limitations in achieving style coherency in scene text editing. Specifically, the generated text style is typically constrained by the typeface used to render the glyph images provided as latent guidance. Based on these observations, we opt for a cross-attention-based approach paired with an inpainting formulation for scene text editing, which enables more precise control over glyph-level features while maintaining style coherence.

\begin{figure}[t]
    \centering
    \includegraphics[width=0.46\textwidth]{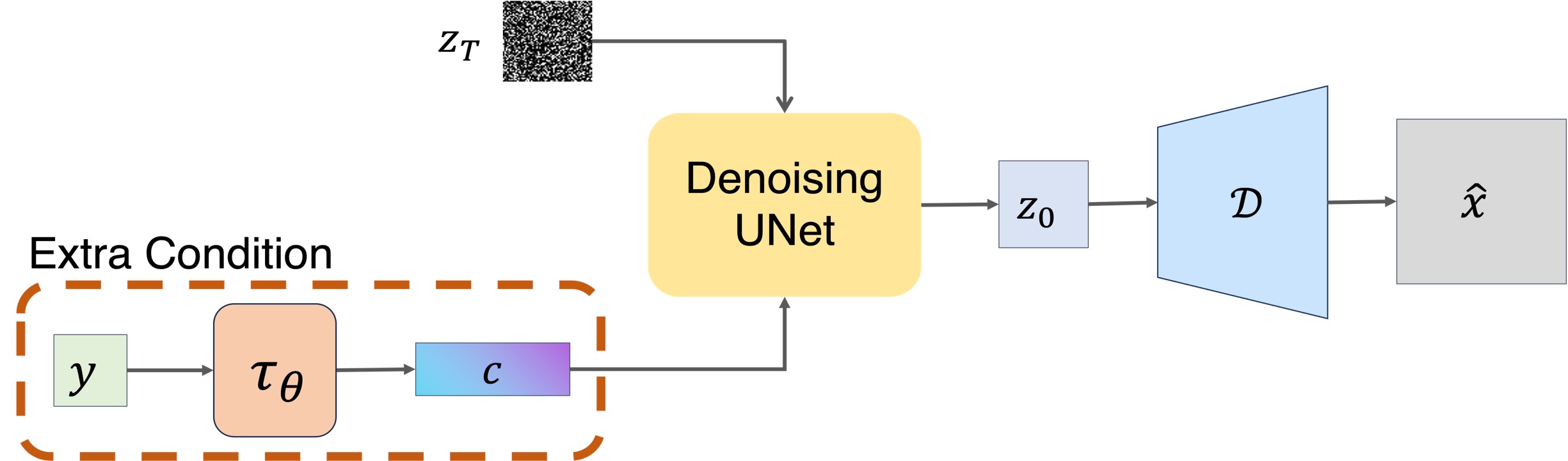}
    \caption{General pipeline for conditioning latent diffusion models with additional guidance signals. An extra condition $y$ is processed through a condition encoder $\tau_\theta$ to produce a condition embedding $c$. This embedding guides the denoising UNet via cross-attention during the iterative denoising process, which transforms the noisy latent $z_T$ into a clean latent representation $z_0$ over $T$ steps. Finally, an image decoder $\mathcal{D}$ converts the latent representation $z_0$ into the final predicted conditioned image $\hat{x}$.}
    \label{fig:condition_gen}
\end{figure}

\begin{figure*}[t]
    \centering
    \includegraphics[width=0.85\textwidth]{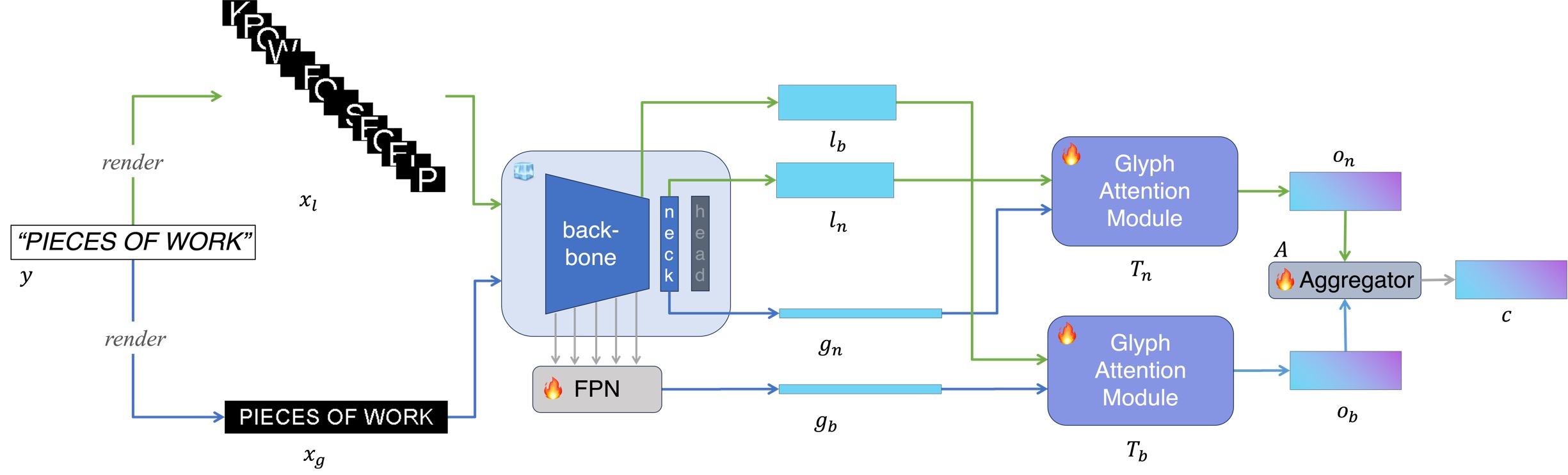}
    \caption{Complete model architecture of \emph{GlyphMastero}. A specialized glyph encoder that introduces stroke-level precise control to the latent diffusion model for scene text editing.}
    \label{fig:combined}
\end{figure*}

\section{Method}

\subsection{Preliminaries}
Diffusion models work by gradually adding noise to data and then learning to reverse this process. The DDPM model~\cite{DBLP:conf/nips/HoJA20} was among the first to demonstrate the effectiveness of diffusion models for image generation, establishing foundational principles that subsequent models have built upon. Latent diffusion models~\cite{DBLP:conf/cvpr/RombachBLEO22} (LDMs) improved upon DDPM by operating in a compressed latent space rather than directly on pixels. They first encode images into a more compact representation, apply the diffusion process,  then decode back to full images. This approach is more computationally efficient than denoising in the pixel space, making it practical for real-world applications. The training objective of LDMs is to minimize:
\begin{equation}
L_{\text{LDM}} := \mathbb{E}_{\mathcal{E}(x), c, \epsilon \sim \mathcal{N}(0,1), t} \left[ \left\| \epsilon - \epsilon_\theta(z_t, c, t) \right\|_2^2 \right]
\label{eq:diffusion}
\end{equation}
where $\mathcal{E}$ is the encoder that compresses $x$ to get latents $z$, $\epsilon$ represents the ground truth noise that is added to $z$ in the forward pass, $\epsilon_\theta$ is the time-dependent model that predicts the noise added given time step $t$, realized by a UNet~\cite{DBLP:conf/miccai/RonnebergerFB15}, and $c$ represents the conditioning embeddings that guide the denoising process with the  cross-attention mechanism. The extra conditioning embeddings $c$ is usually acquired by encoding the extra condition $y$ through a condition encoder $\tau_\theta$ that transforms it into an embedding space, such that $c = \tau_\theta(y)$ (Figure~\ref{fig:condition_gen}). Once the denoising process is finished, the clean latent feature $z_0$ is decoded through a decoder $\mathcal{D}$ to get the image $\hat{x} = \mathcal{D}(z_0)$.

The conditioning mechanism illustrated in Figure~\ref{fig:condition_gen} can be exemplified through the case of text-guided image generation. In this context, the condition $y$ comprises natural language descriptions (or text prompts) which are processed through a $\tau_\theta$ that encodes texts. This encoder is typically implemented using language models, such as CLIP's~\cite{DBLP:conf/aaai/LiLC0LFZ0W23} text encoder or T5~\cite{DBLP:journals/jmlr/RaffelSRLNMZLL20}. The resulting text embedding $c=\tau_\theta(y)$ serves as a guidance signal for the UNet through the cross-attention mechanism. Upon completion of the denoising process, the model yields a refined latent representation $z_0$ that is conditioned on descriptive languages, and is subsequently transformed into the final image through the image decoder, often realized by a VAE~\cite{DBLP:journals/corr/KingmaW13} decoder, to get the generated image $\hat{x} = \mathcal{D}(z_0)$. This cross-attention guided architecture enables the text-to-image diffusion model to learn mappings between arbitrary textual descriptions and their corresponding visual representations during the iterative denoising procedure.

Many practical applications require modifying specific parts of an image rather than generating entirely new ones. While text-to-image diffusion models excel at global image generation, they need to be extended to support such granular control over specific image regions. For text-guided inpainting, Rombach \etal~\cite{DBLP:conf/cvpr/RombachBLEO22} followed LaMa~\cite{DBLP:conf/wacv/SuvorovLMRASKGP22} by introducing additional channels to process a binary mask $m$ that indicates regions to be modified. Specifically, given a binary mask $m$, they first calculate:
\begin{equation}
x_m = x \odot (1-m)
\end{equation}
where $x$ denotes the input image, and $\odot$ denotes the Hadamard (element-wise) product. This operation effectively masks out (sets to zero) the regions marked for inpainting. The latent representation at timestep $t$, denoted as $z_t$, is then augmented by concatenating it with the mask $m$ and the encoded masked image $\mathcal{E}(x_m)$, resulting in:
\begin{equation}
\hat{z}_t = [z_t; m; \mathcal{E}(x_m)]
\label{eq:inpaint}
\end{equation}
where $\mathcal{E}(\cdot)$ is the image encoder function and $[;]$ denotes channel-wise concatenation. The inpainting formulation generalizes text-to-image models by incorporating the binary mask $m$ as an explicit spatial control signal in conjunction with the text condition $c$, enabling the model to preserve contextual fidelity in unmasked regions while performing targeted generation within masked areas.

\subsection{Overall Architecture}
We propose \emph{GlyphMastero}, a novel glyph encoder that produces fine-grained glyph guidance for diffusion models in scene text editing. Relating to the general conditioning pipeline in Figure~\ref{fig:condition_gen}, our approach can be conceptualized as introducing a glyph encoder $\tau_\theta$ to transform text conditions $y$ into fine-grained glyph representations $c$ to guide the denoising UNet to generate scene texts.

In contrast to prior works that directly utilize OCR features for diffusion model guidance - either as an augmentation~\cite{DBLP:journals/corr/abs-2311-03054,DBLP:journals/corr/abs-2304-05568} to or a replacement~\cite{DBLP:conf/iclr/TuoXHGX24} for natural language conditioning - without additional feature processing, our glyph encoder is a learnable and dedicated module that enhances feature representation through hierarchical processing. Our approach generates more fine-grained representations, significantly improving both text accuracy and style preservation in scene text generation.

Following~\cite{DBLP:conf/nips/ChenHL0CW23}, our work builds upon inpainting formulation as described in Eq.\ref{eq:inpaint} to operate at a region-specific granularity, as scene text editing naturally reduces to an inpainting problem where only the masked text regions require generation.

Our \emph{GlyphMastero} is trained jointly with the latent diffusion model using the objective described in Eq.\ref{eq:diffusion}, optimizing the glyph encoder directly through generation-based supervision.

\begin{figure*}[t]
\centering
\includegraphics[width=0.85\textwidth]{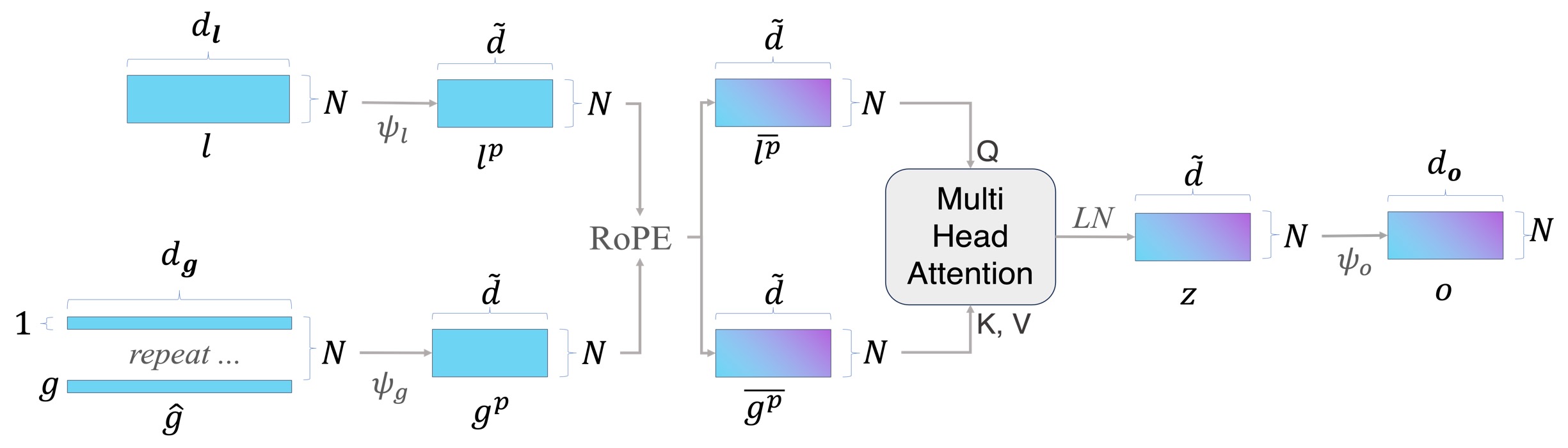} 
\caption{Glyph Attention Module}
\label{fig3}
\end{figure*}

\subsection{Dual-Stream Glyph Integration} \label{sec:dual_stream}
Figure~\ref{fig:combined} illustrates the overall architecture of our glyph encoder. Following~\cite{DBLP:journals/corr/abs-2311-03054}, we incorporate the pretrained PaddleOCR-v4~\cite{paddleocr2023} recognition model - which comprises a backbone, neck, and head - as our feature extractor for glyph images rendered from input texts. We extract two streams of glyph features with the OCR model: a local-level stream (depicted in green arrows) and a global-level stream (depicted in blue arrows), which we systematically integrate through cross-level and multi-scale fusion to derive fine-grained glyph guidance $c$.

In the local-level stream, given a text input $y$, we render a series of single-character glyph images, $x_l \in \mathbb{R}^{N \times H_l \times W_l}$, serving as the local-level representation of the text input. Here, $N$ is the number of characters, and $H_l$ and $W_l$ are the height and width of each single character glyph image, respectively. The OCR model's last-layer backbone output, $l_b$, and the neck output, $l_n$, form the local-level stream feature representations.

In the global-level stream, the input text $y$ is rendered as a unified glyph image $x_g \in \mathbb{R}^{H_g \times W_g}$, where $H_g$ and $W_g$ denote the height and width dimensions, respectively. Similar to the local stream, the neck output feature, $g_n$, is extracted for subsequent processing. Unlike the local stream, we integrate $M$ hierarchical backbone features ($M=5$ in PaddleOCR-v4) for the global stream through a FPN~\cite{DBLP:conf/cvpr/LinDGHHB17}, which fuses high-resolution, fine-grained features in shallow layers with semantic-rich features at lower resolutions in deeper layers, yielding the enhanced backbone features $g_b$.

Having extracted four features: $l_n$ and $l_b$ for the local-level stream, and $g_n$ and $g_b$ for the global-level stream, we then employ two glyph attention modules (Section~\ref{subsec:glyph_trans}) to capture interactions between local and global features for both the backbone and neck features. The cross-level interaction-enhanced features $o_n, o_b \in \mathbb{R}^{N \times d_o}$ are then obtained through two glyph attention modules $T_n$ and $T_b$ as follows:
\begin{equation}
o_n = T_{n}(l_n, g_n), \quad o_b = T_b(l_b, g_b)
\end{equation}
where $d_o$ represents the output dimension of glyph attention modules. We note that $T_n$ and $T_b$ share identical architectural designs; they are independently instantiated to process neck and backbone features respectively.

Finally, an aggregator $A$ concatenates and projects the two features as follows:
\begin{equation}
    c = A(o_b, o_n).
\end{equation}

The resulting condition embedding $c \in \mathbb{R}^{N \times D}$, where $D=d_o$, guides the UNet during both training and inference phases of scene text editing through cross-attention.

\subsection{Glyph Attention Module} \label{subsec:glyph_trans}
The intention of designing the glyph attention module is to use cross-attention to capture the interaction between character-level local features and line-level global features to get a better representation of text glyphs. 

Figure \ref{fig3} illustrates the details of our glyph attention module. Given the local features $l$ and global features $g$ extracted by the OCR model (where this notation applies to both backbone and neck levels, hence omitting the subscripts for notational simplicity). We first repeat the global feature $g \in \mathbb{R}^{1 \times d_g}$ by $N$ times to obtain $\hat{g} \in \mathbb{R}^{N \times d_g}$ to match the local features $l \in \mathbb{R}^{N \times d_{l}}$ with respect to sequence length. Here $N$ is the number of characters in the text input. The features $l$ and $\hat{g}$ are then projected to attention space dimension $\tilde{d}$ through learnable linear transformations $\psi_l$ and $\psi_g$, resulting in projected local and global features $l^p = \psi_l(l)$ and $g^p = \psi_g(\hat{g})$, both in $\mathbb{R}^{N \times \tilde{d}}$.

We add positional embeddings to the two streams of features with rotary positional embedding (RoPE)~\cite{DBLP:journals/corr/abs-2104-09864} by computing:
\begin{equation}
    \bar{l^p}, \bar{g^p} = \text{RoPE}(l^p, g^p)
\end{equation}

To capture interactions between local and global representations, we perform multi-head cross-attention where positionally encoded $\bar{l^p}$ serves as queries and $\bar{g^p}$ as keys and values, followed by a layer normalization (\emph{LN}) to produce the attention map $z \in \mathbb{R}^{N\times \tilde{d}} $. Finally, a linear projection $\psi_o$ is applied to map $z$ from attention dimension $\tilde{d}$ to the output size $d_{o}$ with $o = \psi_o(z) \in \mathbb{R}^{N\times d_o}$.

\begin{figure*}[htbp]
\centering
\begin{tabular}{ccccc}
\emph{Prompt} & \emph{Masked Source} & {DiffUTE} & {AnyText} & \textbf{Ours} \\
\midrule
\raisebox{-.5\height}{\includegraphics{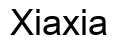}} &
\raisebox{-.5\height}{\includegraphics[width=0.17\textwidth]{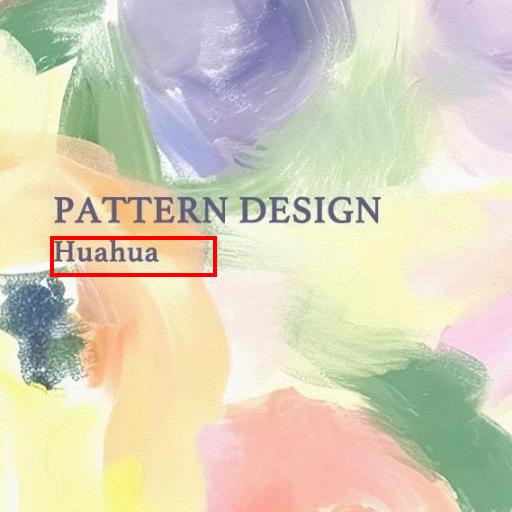}} &
\raisebox{-.5\height}{\includegraphics[width=0.17\textwidth]{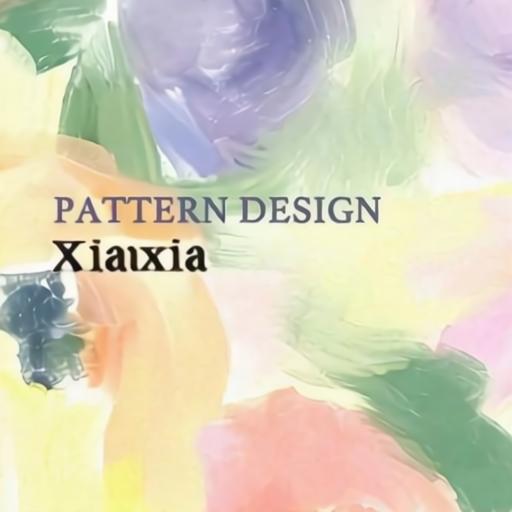}} &
\raisebox{-.5\height}{\includegraphics[width=0.17\textwidth]{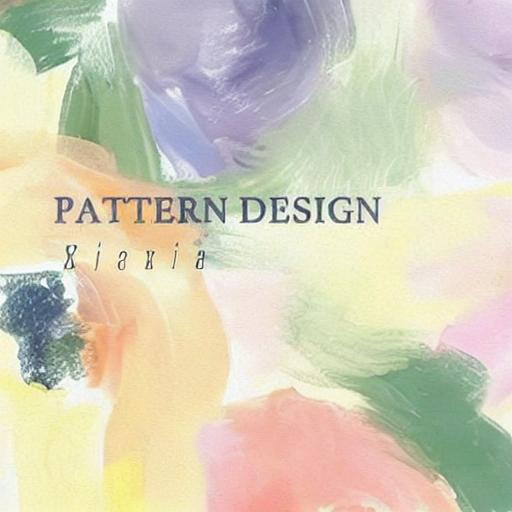}} &
\raisebox{-.5\height}{\includegraphics[width=0.17\textwidth]{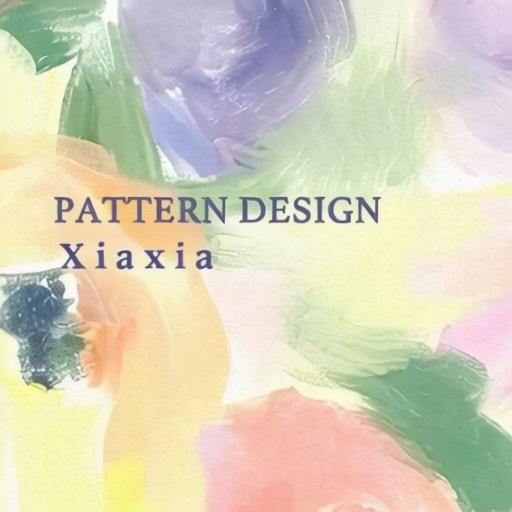}} \\

\raisebox{-.5\height}{\includegraphics{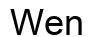}} &
\raisebox{-.5\height}{\includegraphics[width=0.17\textwidth]{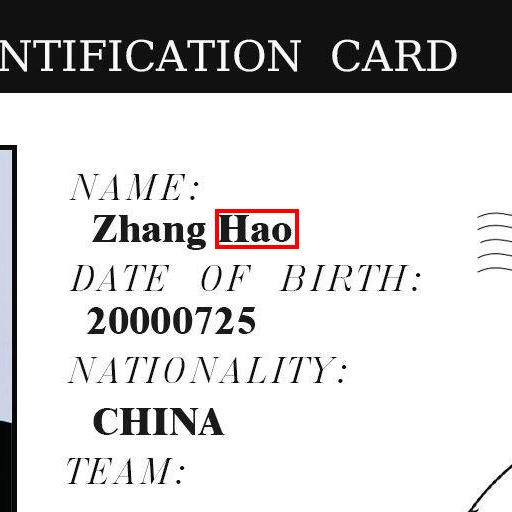}} &
\raisebox{-.5\height}{\includegraphics[width=0.17\textwidth]{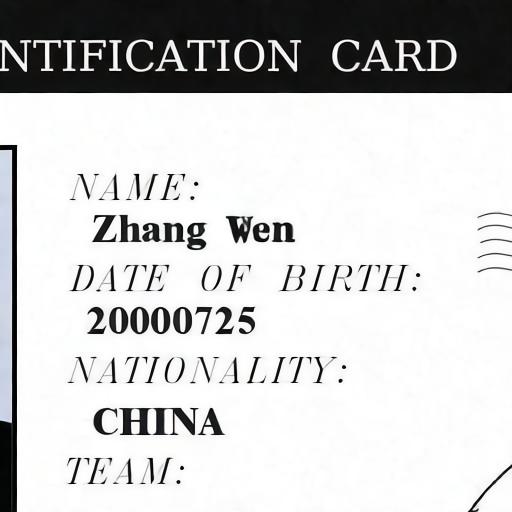}} &
\raisebox{-.5\height}{\includegraphics[width=0.17\textwidth]{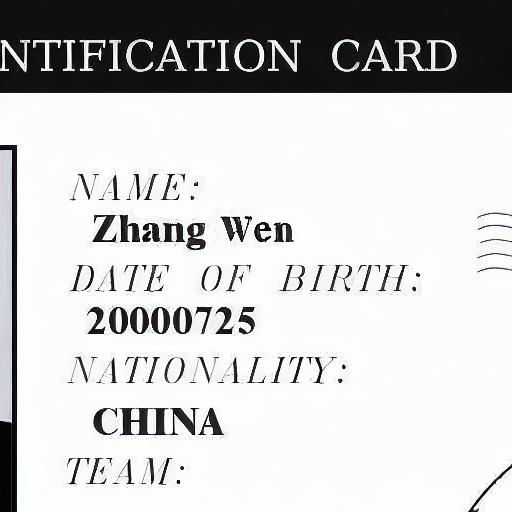}} &
\raisebox{-.5\height}{\includegraphics[width=0.17\textwidth]{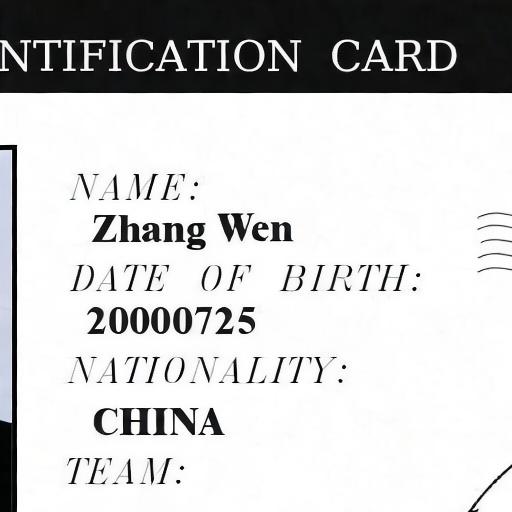}} \\

\raisebox{-.5\height}{\includegraphics{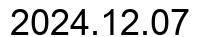}} &
\raisebox{-.5\height}{\includegraphics[width=0.17\textwidth]{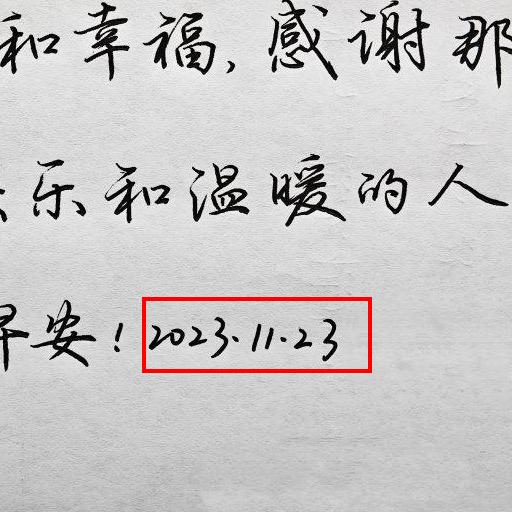}} &
\raisebox{-.5\height}{\includegraphics[width=0.17\textwidth]{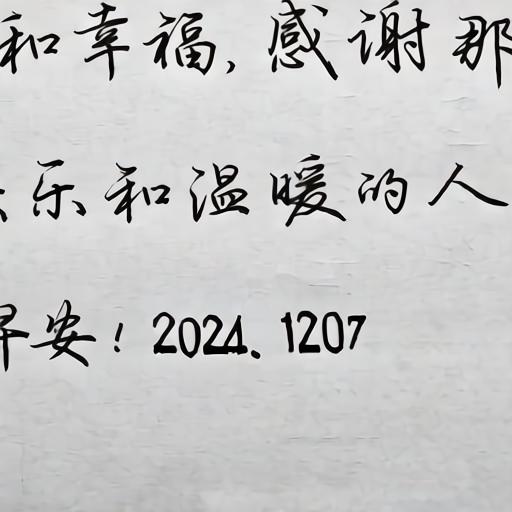}} &
\raisebox{-.5\height}{\includegraphics[width=0.17\textwidth]{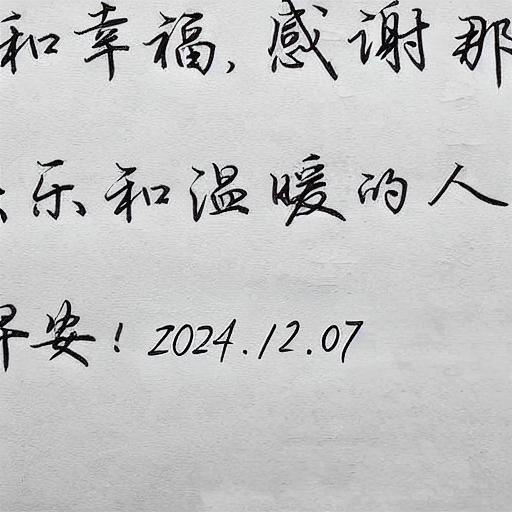}} &
\raisebox{-.5\height}{\includegraphics[width=0.17\textwidth]{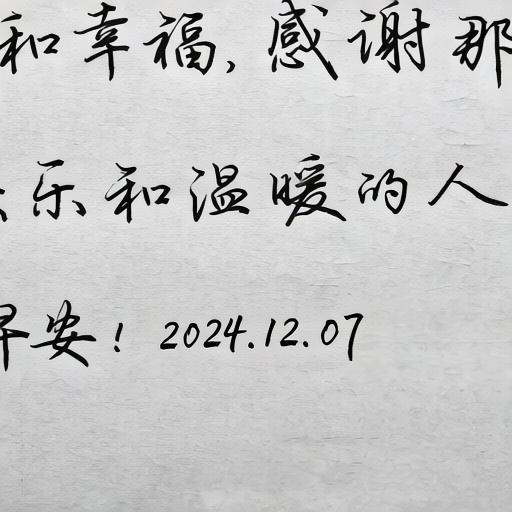}} \\

\raisebox{-.5\height}{\includegraphics{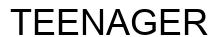}} &
\raisebox{-.5\height}{\includegraphics[width=0.17\textwidth]{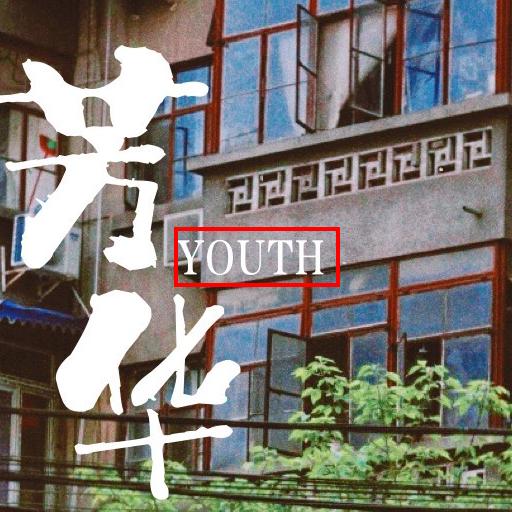}} &
\raisebox{-.5\height}{\includegraphics[width=0.17\textwidth]{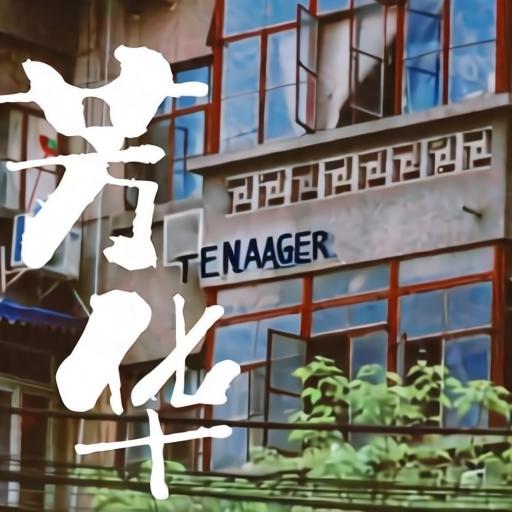}} &
\raisebox{-.5\height}{\includegraphics[width=0.17\textwidth]{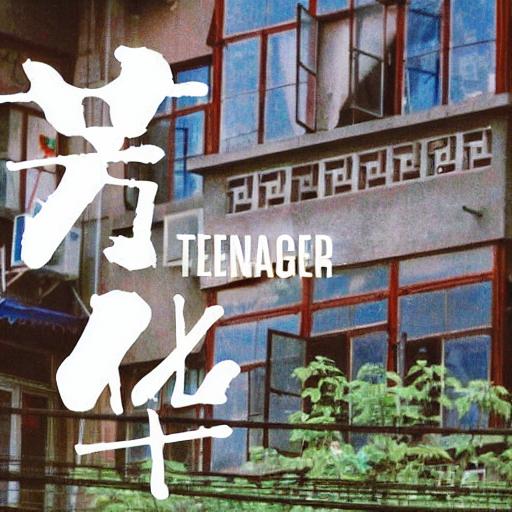}} &
\raisebox{-.5\height}{\includegraphics[width=0.17\textwidth]{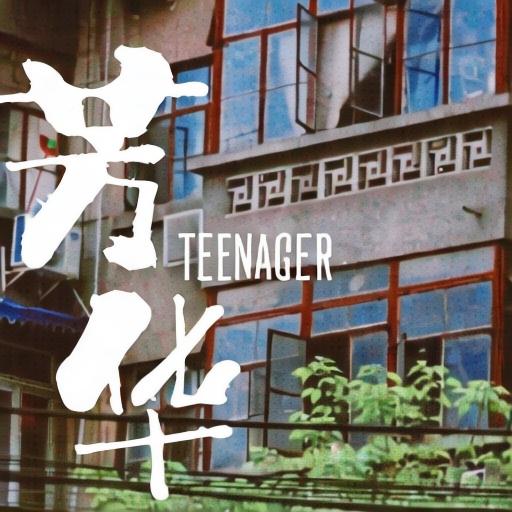}} \\

\raisebox{-.5\height}{
\shortstack{\includegraphics{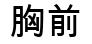} \\ \textit{(in front of chest)}}
} &
\raisebox{-.5\height}{\includegraphics[width=0.17\textwidth]{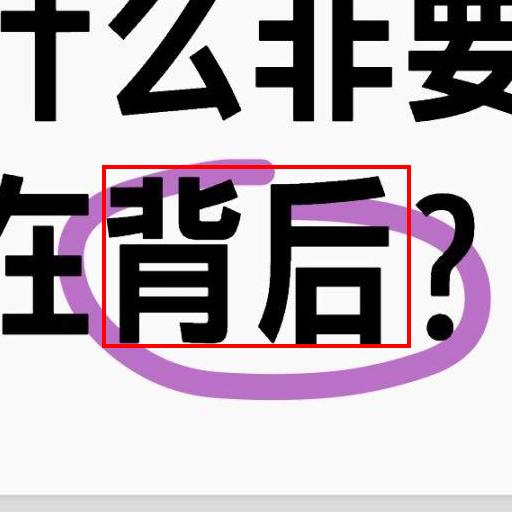}} &
\raisebox{-.5\height}{\includegraphics[width=0.17\textwidth]{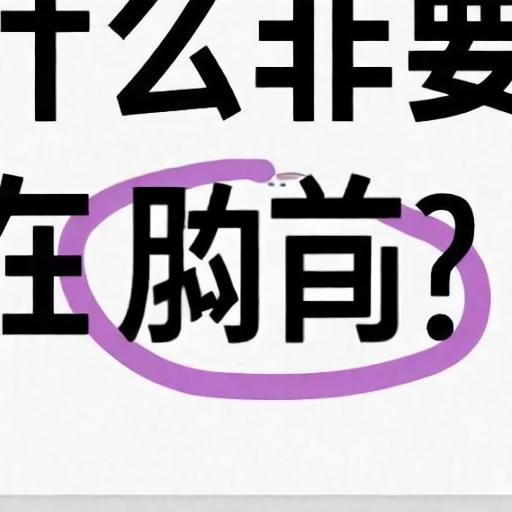}} &
\raisebox{-.5\height}{\includegraphics[width=0.17\textwidth]{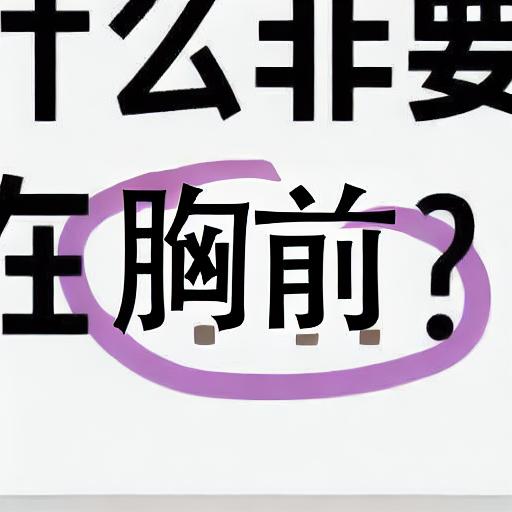}} &
\raisebox{-.5\height}{\includegraphics[width=0.17\textwidth]{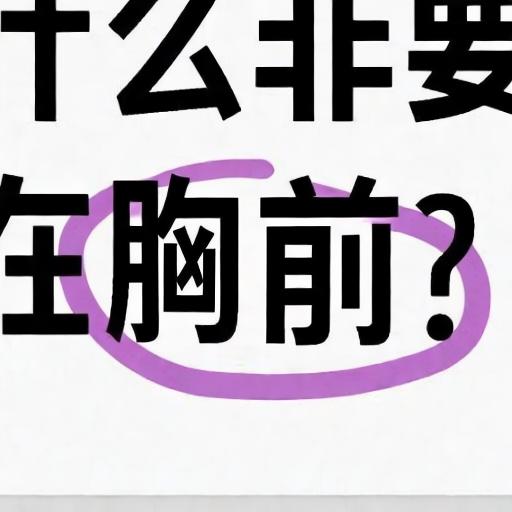}} \\

\raisebox{-.5\height}{
\shortstack{\includegraphics{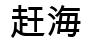} \\ \textit{(beachcombing)}}
} &
\raisebox{-.5\height}{\includegraphics[width=0.17\textwidth]{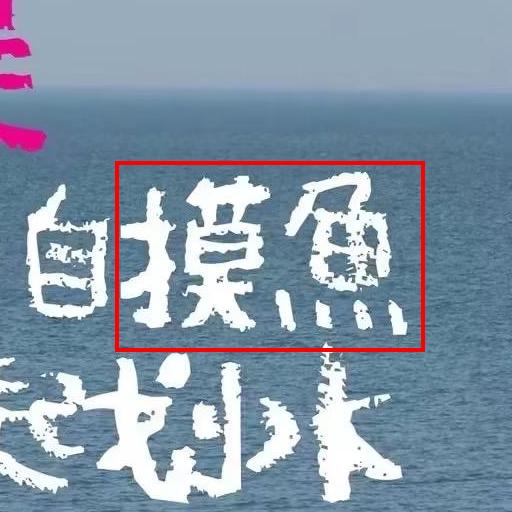}} &
\raisebox{-.5\height}{\includegraphics[width=0.17\textwidth]{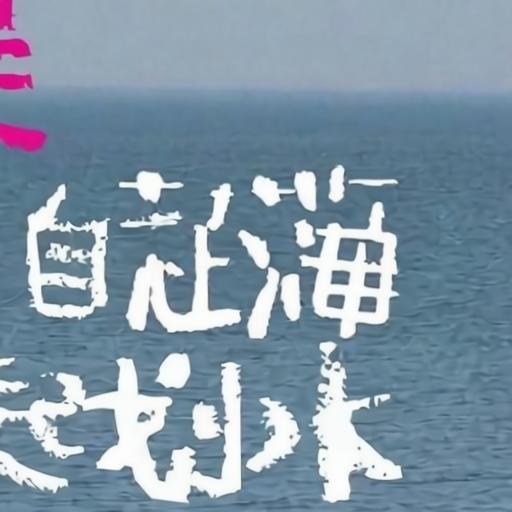}} &
\raisebox{-.5\height}{\includegraphics[width=0.17\textwidth]{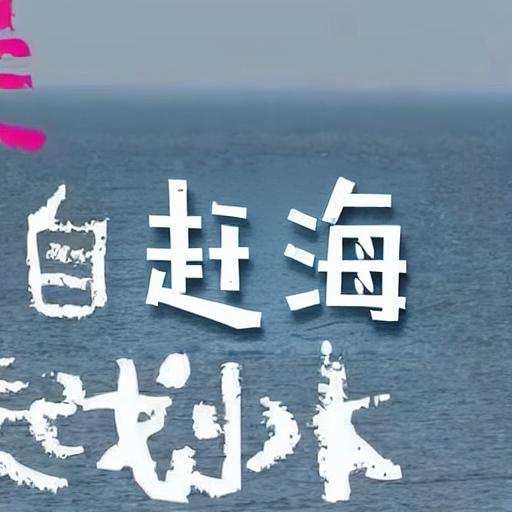}} &
\raisebox{-.5\height}{\includegraphics[width=0.17\textwidth]{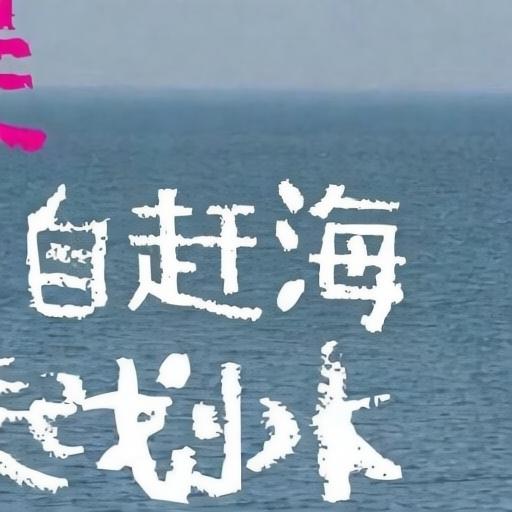}} \\

\raisebox{-.5\height}{
\shortstack{\includegraphics{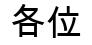} \\ \textit{(everybody)}}
} &
\raisebox{-.5\height}{\includegraphics[width=0.17\textwidth]{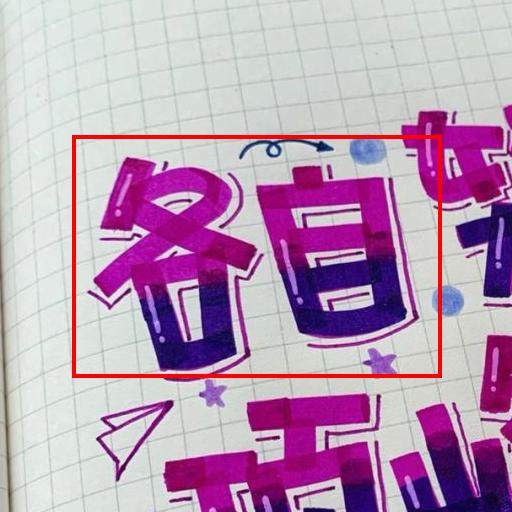}} &
\raisebox{-.5\height}{\includegraphics[width=0.17\textwidth]{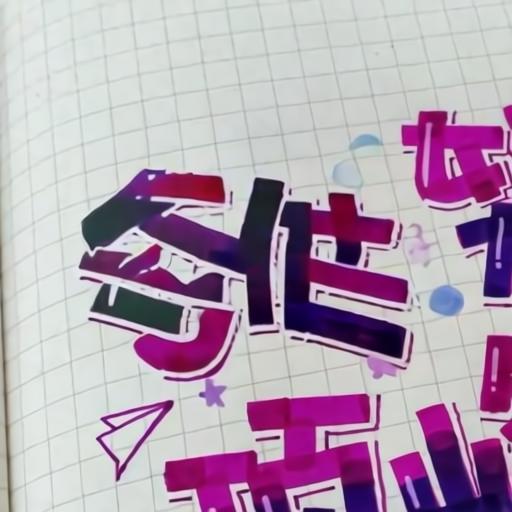}} &
\raisebox{-.5\height}{\includegraphics[width=0.17\textwidth]{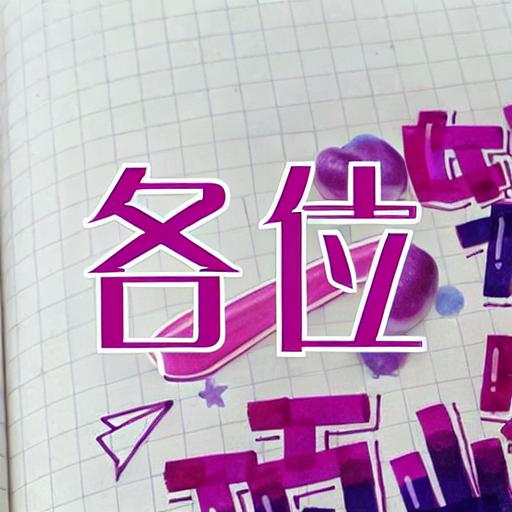}} &
\raisebox{-.5\height}{\includegraphics[width=0.17\textwidth]{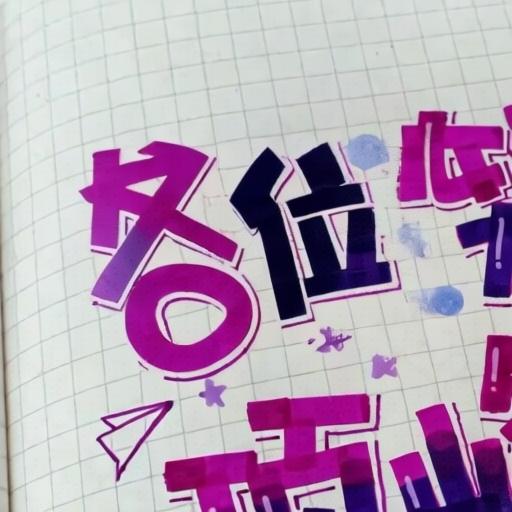}}
\end{tabular}
\caption{Qualitative comparison of scene text editing methods. Our GlyphMastero framework demonstrates superior text style preservation and content replacement accuracy. For Chinese cases (with English translations in brackets), our method achieves more precise text generation than DiffUTE and better style preservation than AnyText.}
\label{fig:method_comparison}
\end{figure*}

\section{Experiments}

\begin{table*}[h]
\centering
\small
\setlength{\tabcolsep}{3pt}
\begin{tabular}{@{}l|cccc|cccc@{}}
\toprule
\multirow{3}{*}{Model} & \multicolumn{4}{c|}{Text Accuracy} & \multicolumn{4}{c}{Style Similarity} \\
\cmidrule(lr){2-5} \cmidrule(lr){6-9}
 & \multicolumn{2}{c}{Sen.Acc $\uparrow$} & \multicolumn{2}{c|}{CER $\downarrow$} & \multicolumn{2}{c}{FID $\downarrow$} & \multicolumn{2}{c}{LPIPS $\downarrow$} \\
& English & Chinese & English & Chinese & English & Chinese & English & Chinese \\
\midrule
DiffUTE & 0.3319 & 0.2523 & 0.3186 & 0.4048 & 14.3176 & 24.9295 & 0.1313 & 0.2056 \\
AnyText & 0.6067 & 0.5801 & 0.1730 & 0.2088 & 10.4257 & 24.9004 & 0.1098 & 0.1978 \\
Ours & \textbf{0.8170} & \textbf{0.7301} & \textbf{0.0741} & \textbf{0.1341} & \textbf{4.6101} & \textbf{11.8915} & \textbf{0.0545} & \textbf{0.1007} \\
\bottomrule
\end{tabular}
\caption{Comparison with state-of-the-art multi-lingual (English and Chines) methods. $\uparrow$/$\downarrow$ indicates higher/lower is better. Our method outperforms previous works across all metrics.}
\label{tab:model_comparison}
\end{table*}

\subsection{Dataset}
For both training and experimental evaluation, we utilize AnyWord-3M~\cite{DBLP:conf/iclr/TuoXHGX24}, a comprehensive dataset designed for scene text generation and editing. We utilize the dataset specifically for scene text editing scanario: each sample is an image containing multiple line-level annotations, which consists of text content, and its corresponding polygonal position as target area to generate new text. Since each image can have multiple text regions, we randomly sample one text-position pair in training. Our dataset formulation can be written as:
\begin{equation}
    \mathcal{D} = \{(I_i, \{(T_{ij}, P_{ij})\}_{j=1}^{M_i})\}_{i=1}^N
\end{equation}
where $N$ is the total number of images, approximately 3.5 million in case of AnyWord dataset. $M_i$ is number of text-position pairs in the $i$th image. Hence $(T_{ij}, P_{ij})$ is a single data sample, where $T_{ij}$ represents the target text string and $P_{ij}$ denotes its spatial location encoded as a quadrilateral polygon with four vertices. During training and evaluation, a binary mask is generated based on $T_{ij}$ to delineate the text region for inpainting.

We evaluate our model using AnyText-Eval~\cite{DBLP:conf/iclr/TuoXHGX24} for quantitative analysis (2000 images, 4181 English and 2092 Chinese text-position pairs) and a curated dataset of 80 challenging stylistic images (120 text-position pairs) for qualitative assessment. While AnyText-Eval provides statistical coverage, its target texts match the original, not fully reflecting real-world editing tasks. Our curated dataset focuses on challenging stylistic variations and incorporates new text contents. 

\subsection{Baseline Methods}
For multi-lingual editing comparisons, we benchmark against state-of-the-art models DiffUTE~\cite{DBLP:conf/nips/ChenXGLZLMZW23} and AnyText~\cite{DBLP:journals/corr/abs-2311-03054}. For AnyText, we use the public AnyWord-3M checkpoint. We trained DiffUTE on AnyWord-3M for 15 epochs with a batch size of 256 and null condition probability of 0.1, matching our model's hyperparameters. 

Many existing methods primarily focused on English
text, numbers, and punctuation editing. For English-only comparisons, we evaluated SRNet~\cite{DBLP:conf/mm/WuZLHLDB19}, MOSTEL~\cite{DBLP:conf/aaai/QuTXXW023}, DiffSTE~\cite{DBLP:journals/corr/abs-2304-05568}, TextDiffuser~\cite{DBLP:conf/nips/ChenHL0CW23}, and TextCtrl~\cite{DBLP:conf/nips/ZengSLYZ24} using their publicly available checkpoints. 

\subsection{Implementation Details} 
We initialize the UNet with Stable-Diffusion 2.1 inpainting weights~\cite{DBLP:conf/cvpr/RombachBLEO22}. And the pretrained OCR recognition model is initialized with PaddleOCR-v4~\cite{paddleocr2023}.

Each glyph attention module has one multi-head attention layer with $4$ heads, with model size $\tilde{d} = 512$ and $d_{o} = 1024$. Their weights are initialized by Xavier initialization~\cite{DBLP:journals/corr/abs-1907-04108}. 

We implemented the FPN specifically suited for the PaddleOCR-V4's backbone module. The module processes a hierarchical feature pyramid ${x_1, x_2, x_3, x_4, x_5}$, where each feature map $x_i \in \mathbb{R}^{C_i \times H_i \times W_i}$ represents different levels of glyph features. Lateral connections first project all input features to a common hidden dimension $D$ through 1×1 convolutions: $f_i: \mathbb{R}^{C_i \times H_i \times W_i} \rightarrow \mathbb{R}^{D \times H_i \times W_i}$. Starting from $p_5 = c_5$, for each level $i = 4, 3, 2, 1$, features are then fused as:
\begin{equation}
p_i = g_i(u(p_{i+1}) + c_i)
\end{equation}
where $u(\cdot)$ denotes bilinear upsampling, $g_i$ is a 3×3 convolution, and $c_i$ represents lateral features. The final feature map undergoes channel projection, downsampling, and pooling to produce the enhanced global feature $g_b$ with dimensions matching $x_5$.

The guidance aggregator consists of two parallel linear projection branches that reduce the dimensionality of input embeddings from $1024$ to $512$ dimensions, then concatenating them back to $1024$ dimensions.

Our model is trained for 15 epochs with a global batch size of 256 on 8 V100S-32G GPUs. We also use a null condition probability of 0.1 to allow classifier-free guidance (CFG) in inference. See the \emph{Supplementary Material} for our analysis on effects of CFG in scene text editing task.

In inference, all methods use DDIM~\cite{DBLP:conf/iclr/SongME21} sampler with 20 steps denoising steps, with CFG scale 9 for AnyText and 3 for ours and DiffUTE.

\begin{table}[h]
\centering
\setlength{\tabcolsep}{3pt}
\footnotesize
\begin{tabular}{l|cc|cc}
\hline
\multirow{2}{*}{Methods} & \multicolumn{2}{c|}{Accuracy} & \multicolumn{2}{c}{Similarity} \\
\cline{2-5}
& Sen.Acc$\uparrow$ & CER$\downarrow$ & FID$\downarrow$ & LPIPS$\downarrow$ \\
\hline
SRNet & 0.3994 & 0.3802 & 27.00 & 0.1073 \\
MOSTEL & 0.5781 & 0.1984 & 50.24 & 0.4883 \\
DiffSTE & 0.5120 & 0.1993 & 14.06 & 0.1543 \\
TextDiffuser & 0.5260 & 0.2350 & 33.15 & 0.2406 \\
DiffUTE & 0.3319 & 0.3186 & 14.32 & 0.1313 \\
AnyText & 0.6067 & 0.1730 & 10.43 & 0.1098 \\
TextCtrl & 0.7654 & 0.0940 & \textbf{4.15} & \textbf{0.0425} \\
Ours & \textbf{0.8170} & \textbf{0.0741} & 4.61 & 0.0545 \\
\hline
\end{tabular}
\caption{English only quantitative comparison results.}
\label{tab:textctrl_cmp}
\vspace{-2em}
\end{table}

\begin{table*}[t]
\centering
\small
\setlength{\tabcolsep}{3pt}
\begin{tabular}{@{}l|cccc|cccc@{}}
\toprule
\multirow{3}{*}{Component} & \multicolumn{4}{c|}{Text Accuracy} & \multicolumn{4}{c}{Style Similarity} \\
\cmidrule(lr){2-5} \cmidrule(lr){6-9}
 & \multicolumn{2}{c}{Sen.Acc $\uparrow$} & \multicolumn{2}{c|}{CER $\downarrow$} & \multicolumn{2}{c}{FID $\downarrow$} & \multicolumn{2}{c}{LPIPS $\downarrow$} \\
& English & Chinese & English & Chinese & English & Chinese & English & Chinese \\
\midrule
Full model & \bf 0.5494 & \bf 0.5120 & \bf 0.1766 & \bf 0.2367 & 30.9095 & 51.3762 & \bf 0.1190 & 0.2134 \\
\hline
$-$ FPN & 0.4536 & 0.3698 & 0.2470 & 0.3314 & 32.4550 & 53.0127 & 0.1247 & 0.2208 \\
$-$ $T_b$ & 0.5065 & 0.4271 & 0.2128 & 0.2987 & \bf 30.5866 & \bf 49.6588 & 0.1196 & \bf 0.2110 \\
$-$ $T_n$ \hspace{0.3em} \emph{w/ $l_n$} & 0.3263 & 0.2735 & 0.3137 & 0.3916 & 30.6271 & 51.6288 & 0.1211 & 0.2121 \\
\hspace*{2.7em} \emph{w/ $g_n$} & 0.1003 & 0.0719 & 0.5727 & 0.6579 & 34.7661 & 51.5788 & 0.1412 & 0.2334 \\
\bottomrule
\end{tabular}
\caption{Ablation study results. When removing $T_n$, we experiment with two scenarios: using local features ($l_n$) and global features ($g_n$).}
\label{tab:ablation_results}
\end{table*}

\subsection{Quantitative Comparison} 
We measure four metrics for different methods: sentence accuracy (Sen.Acc) and character error rate (CER) for text content fidelity; Fréchet Inception Distance (FID)~\cite{Seitzer2020FID} and LPIPS~\cite{zhang2018perceptual} for style similarity. Sen.Acc measures line-level accuracy, while CER is for character-level accuracy. FID measures distribution-level style similarity, while LPIPS focuses on sample-level similarity. We average LPIPS distances over all samples for the final measurement. FID and LPIPS are measured between cropped ground truth images and generated text regions.

Table \ref{tab:model_comparison} shows our comparisons with prior arts that has multi-lingual scene text editing capbility. It is seen that our method significantly outperforms prior arts in both text generation accuracy and style similarity. Our overall sentence accuracy (averaging English and Chinese results) is 48.14\% and 18.02\% higher than DiffUTE and AnyText, respectively; CER is 25.76\% and 8.68\% lower. For style similarity, we also achieved with substantially lower FID and LPIPS distances compared to DiffUTE and AnyText. For FID evaluations, our method's distance is 57.95\% lower than DiffUTE, and 53.28\% lower than AnyText, on average. Our LPIPS distance is 53.93\% lower than DiffUTE, and 49.54\% lower than AnyText.

Table~\ref{tab:textctrl_cmp} presents English-only comparison results. Our method achieved the highest accuracy metrics while outperforming all other methods in similarity measures except TextCtrl, which showed marginally better FID and LPIPS. We also evaluated our approach on TextCtrl's ScenePair dataset (1,285 test cases), using their GitHub-published results and evaluation scripts, also achieving favorable outcomes. Full comparison results are available in the \emph{Supplementary Material}.

\subsection{Qualitative Comparison}
Figure \ref{fig:method_comparison} presents comparative results among our method, AnyText, and DiffUTE. Our approach demonstrates substantial improvements over prior arts in both text accuracy and style preservation. We specifically include Chinese characters in our evaluation as they present a more rigorous test case, featuring complex glyph structures with multiple strokes. The superior performance in accurately generating these intricate characters demonstrates our GlyphMastero's effectiveness in capturing stroke-level precision features. More extensive qualitative results are available in the \emph{Supplementary Material}.

\subsection{Ablation Studies}
To evaluate the effectiveness of individual components, we conducted a systematic ablation study on GlyphMastero. The experiments are performed using a subset of 375K images from the Anyword-3M dataset, with training conducted for 15 epochs using a batch size of 256 on four NVIDIA A100-40GB GPUs. Ablation study results are recorded in Table \ref{tab:ablation_results}.

First, we removed the FPN that fuses multi-scale backbone feature maps, and replaced it with the final backbone layer output in global-level stream. The model trained with this modification led to a 22.42\% drop in average sentence accuracy for English and Chinese texts and a 28.54\% increase in CER (Table \ref{tab:ablation_results}). These results demonstrate that fusing high-resolution features from shallow layers with semantic-rich features from deeper layers significantly impacts the generated text quality.

Next, we removed the backbone glyph attention module $T_b$, retaining only the neck glyph attention module $T_n$ for cross-level feature capture. This resulted in a 13.68\% drop in average sentence accuracy and a 19.20\% increase in CER (Table \ref{tab:ablation_results}) compare to the full model. Notably, removing both FPN and $T_b$ yielded better performance than removing FPN alone. We attribute this to the FPN's role in adaptively refining features for $T_b$'s cross-level feature capture. Without FPN, $T_b$ struggles to effectively process directly extracted backbone features.

Finally, we removed $T_n$ and used only neck features as guidance, testing two scenarios. First, with position-embedded local neck features $l_n$ (\textit{w/} $l_n$), we observed a 43.49\% drop in average sentence accuracy and 41.40\% increase in CER. These significant degradations highlight the crucial role of our glyph attention module in capturing local-global feature relationships. Second, using global neck features $g_n$ as prior guidance (\textit{w/} $g_n$) - equivalent to AnyText's OCR feature utilization - yielded the poorest results. We attribute this to $g_n$ prior reducing embedding length from $N$ to $1$, while local neck features $l_n$ maintained length $N$.

The style similarity metrics (FID and LPIPS) remain consistent throughout ablation experiments, with only slight variations. This stability aligns with GlyphMastero's core function as a glyph encoder, where its components focus on providing fine-grained glyph guidance, while style preservation is primarily handled by the latent diffusion inpainting model's inherent capacity for maintaining local visual consistency. 

\section{Conclusion \& Limitations}
We present \emph{GlyphMastero}, a novel trainable glyph encoder that advances scene text editing through effective modeling of hierarchical relationships between local character-level features and global text-line structures. Our approach addresses the challenge of encoding complex glyph structures through multi-scale feature incorporation and explicit modeling of hierarchical glyph relationships. The effectiveness of our approach is demonstrated by an 18.02\% improvement in sentence accuracy compare to the previous state-of-the-art method and significant enhancement in style preservation, with text-region FID reduced by 53.28\%.

One limitation of our approach is that accuracy for generating long text, though improved over prior work, still lags behind shorter text. We attribute this to training data limited to $512 \times 512$ resolution and constraints of the base latent diffusion model. Future work will explore higher-resolution training and more robust base models to address this limitation.

\section{Acknowledgement}
Xiaolin Hu was supported by the National Natural Science Foundation of China (No. U2341228).

{
    \small
    \bibliographystyle{ieeenat_fullname}
    \bibliography{main}

\begin{thebibliography}{31}
\providecommand{\natexlab}[1]{#1}
\providecommand{\url}[1]{\texttt{#1}}
\expandafter\ifx\csname urlstyle\endcsname\relax
  \providecommand{\doi}[1]{doi: #1}\else
  \providecommand{\doi}{doi: \begingroup \urlstyle{rm}\Url}\fi

\bibitem[Chen et~al.(2023{\natexlab{a}})Chen, Xu, Gu, Lan, Zheng, Li, Meng, Zhu, and Wang]{DBLP:conf/nips/ChenXGLZLMZW23}
Haoxing Chen, Zhuoer Xu, Zhangxuan Gu, Jun Lan, Xing Zheng, Yaohui Li, Changhua Meng, Huijia Zhu, and Weiqiang Wang.
\newblock Diffute: Universal text editing diffusion model.
\newblock In \emph{Advances in Neural Information Processing Systems 36: Annual Conference on Neural Information Processing Systems 2023, NeurIPS 2023, New Orleans, LA, USA, December 10 - 16, 2023}, 2023{\natexlab{a}}.

\bibitem[Chen et~al.(2023{\natexlab{b}})Chen, Huang, Lv, Cui, Chen, and Wei]{DBLP:conf/nips/ChenHL0CW23}
Jingye Chen, Yupan Huang, Tengchao Lv, Lei Cui, Qifeng Chen, and Furu Wei.
\newblock Textdiffuser: Diffusion models as text painters.
\newblock In \emph{Advances in Neural Information Processing Systems 36: Annual Conference on Neural Information Processing Systems 2023, NeurIPS 2023, New Orleans, LA, USA, December 10 - 16, 2023}, 2023{\natexlab{b}}.

\bibitem[Goodfellow et~al.(2014)Goodfellow, Pouget{-}Abadie, Mirza, Xu, Warde{-}Farley, Ozair, Courville, and Bengio]{DBLP:conf/nips/GoodfellowPMXWOCB14}
Ian~J. Goodfellow, Jean Pouget{-}Abadie, Mehdi Mirza, Bing Xu, David Warde{-}Farley, Sherjil Ozair, Aaron~C. Courville, and Yoshua Bengio.
\newblock Generative adversarial nets.
\newblock In \emph{Advances in Neural Information Processing Systems 27: Annual Conference on Neural Information Processing Systems 2014, December 8-13 2014, Montreal, Quebec, Canada}, pages 2672--2680, 2014.

\bibitem[Ho et~al.(2020)Ho, Jain, and Abbeel]{DBLP:conf/nips/HoJA20}
Jonathan Ho, Ajay Jain, and Pieter Abbeel.
\newblock Denoising diffusion probabilistic models.
\newblock In \emph{Advances in Neural Information Processing Systems 33: Annual Conference on Neural Information Processing Systems 2020, NeurIPS 2020, December 6-12, 2020, virtual}, 2020.

\bibitem[Huang et~al.(2022)Huang, Fu, Zhang, and Qiao]{DBLP:journals/corr/abs-2207-09649}
Qirui Huang, Bin Fu, Aozhong Zhang, and Yu Qiao.
\newblock Gentext: Unsupervised artistic text generation via decoupled font and texture manipulation.
\newblock \emph{CoRR}, abs/2207.09649, 2022.

\bibitem[Ji et~al.(2023)Ji, Zhang, Wang, Hou, Zhang, Price, and Chang]{DBLP:journals/corr/abs-2304-05568}
Jiabao Ji, Guanhua Zhang, Zhaowen Wang, Bairu Hou, Zhifei Zhang, Brian Price, and Shiyu Chang.
\newblock Improving diffusion models for scene text editing with dual encoders.
\newblock \emph{CoRR}, abs/2304.05568, 2023.

\bibitem[Kingma and Welling(2014)]{DBLP:journals/corr/KingmaW13}
Diederik~P. Kingma and Max Welling.
\newblock Auto-encoding variational bayes.
\newblock In \emph{2nd International Conference on Learning Representations, {ICLR} 2014, Banff, AB, Canada, April 14-16, 2014, Conference Track Proceedings}, 2014.

\bibitem[Lee et~al.(2021)Lee, Kim, Kim, Yim, Shin, Lee, and Park]{DBLP:journals/corr/abs-2107-11041}
Junyeop Lee, Yoonsik Kim, Seonghyeon Kim, Moonbin Yim, Seung Shin, Gayoung Lee, and Sungrae Park.
\newblock Rewritenet: Realistic scene text image generation via editing text in real-world image.
\newblock \emph{CoRR}, abs/2107.11041, 2021.

\bibitem[Li et~al.(2023)Li, Lv, Chen, Cui, Lu, Flor{\^{e}}ncio, Zhang, Li, and Wei]{DBLP:conf/aaai/LiLC0LFZ0W23}
Minghao Li, Tengchao Lv, Jingye Chen, Lei Cui, Yijuan Lu, Dinei A.~F. Flor{\^{e}}ncio, Cha Zhang, Zhoujun Li, and Furu Wei.
\newblock Trocr: Transformer-based optical character recognition with pre-trained models.
\newblock In \emph{Thirty-Seventh {AAAI} Conference on Artificial Intelligence, {AAAI} 2023, Thirty-Fifth Conference on Innovative Applications of Artificial Intelligence, {IAAI} 2023, Thirteenth Symposium on Educational Advances in Artificial Intelligence, {EAAI} 2023, Washington, DC, USA, February 7-14, 2023}, pages 13094--13102. {AAAI} Press, 2023.

\bibitem[Lin et~al.(2017)Lin, Doll{\'{a}}r, Girshick, He, Hariharan, and Belongie]{DBLP:conf/cvpr/LinDGHHB17}
Tsung{-}Yi Lin, Piotr Doll{\'{a}}r, Ross~B. Girshick, Kaiming He, Bharath Hariharan, and Serge~J. Belongie.
\newblock Feature pyramid networks for object detection.
\newblock In \emph{2017 {IEEE} Conference on Computer Vision and Pattern Recognition, {CVPR} 2017, Honolulu, HI, USA, July 21-26, 2017}, pages 936--944. {IEEE} Computer Society, 2017.

\bibitem[Ma et~al.(2023)Ma, Zhao, Chen, Wang, Niu, Lu, and Lin]{DBLP:journals/corr/abs-2303-17870}
Jian Ma, Mingjun Zhao, Chen Chen, Ruichen Wang, Di Niu, Haonan Lu, and Xiaodong Lin.
\newblock Glyphdraw: Learning to draw chinese characters in image synthesis models coherently.
\newblock \emph{CoRR}, abs/2303.17870, 2023.

\bibitem[{PaddlePaddle}(2023)]{paddleocr2023}
{PaddlePaddle}.
\newblock {PP-OCRv4}.
\newblock \url{https://github.com/PaddlePaddle/PaddleOCR/blob/release/2.7/doc/doc_ch/PP-OCRv4_introduction.md}, 2023.
\newblock Accessed: 2024-07-31.

\bibitem[Qu et~al.(2023)Qu, Tan, Xie, Xu, Wang, and Zhang]{DBLP:conf/aaai/QuTXXW023}
Yadong Qu, Qingfeng Tan, Hongtao Xie, Jianjun Xu, YuXin Wang, and Yongdong Zhang.
\newblock Exploring stroke-level modifications for scene text editing.
\newblock In \emph{Thirty-Seventh {AAAI} Conference on Artificial Intelligence, {AAAI} 2023, Thirty-Fifth Conference on Innovative Applications of Artificial Intelligence, {IAAI} 2023, Thirteenth Symposium on Educational Advances in Artificial Intelligence, {EAAI} 2023, Washington, DC, USA, February 7-14, 2023}, pages 2119--2127. {AAAI} Press, 2023.

\bibitem[Radford et~al.(2021)Radford, Kim, Hallacy, Ramesh, Goh, Agarwal, Sastry, Askell, Mishkin, Clark, Krueger, and Sutskever]{DBLP:conf/icml/RadfordKHRGASAM21}
Alec Radford, Jong~Wook Kim, Chris Hallacy, Aditya Ramesh, Gabriel Goh, Sandhini Agarwal, Girish Sastry, Amanda Askell, Pamela Mishkin, Jack Clark, Gretchen Krueger, and Ilya Sutskever.
\newblock Learning transferable visual models from natural language supervision.
\newblock In \emph{Proceedings of the 38th International Conference on Machine Learning, {ICML} 2021, 18-24 July 2021, Virtual Event}, pages 8748--8763. {PMLR}, 2021.

\bibitem[Raffel et~al.(2020)Raffel, Shazeer, Roberts, Lee, Narang, Matena, Zhou, Li, and Liu]{DBLP:journals/jmlr/RaffelSRLNMZLL20}
Colin Raffel, Noam Shazeer, Adam Roberts, Katherine Lee, Sharan Narang, Michael Matena, Yanqi Zhou, Wei Li, and Peter~J. Liu.
\newblock Exploring the limits of transfer learning with a unified text-to-text transformer.
\newblock \emph{J. Mach. Learn. Res.}, 21:\penalty0 140:1--140:67, 2020.

\bibitem[Rombach et~al.(2022)Rombach, Blattmann, Lorenz, Esser, and Ommer]{DBLP:conf/cvpr/RombachBLEO22}
Robin Rombach, Andreas Blattmann, Dominik Lorenz, Patrick Esser, and Bj{\"{o}}rn Ommer.
\newblock High-resolution image synthesis with latent diffusion models.
\newblock In \emph{{IEEE/CVF} Conference on Computer Vision and Pattern Recognition, {CVPR} 2022, New Orleans, LA, USA, June 18-24, 2022}, pages 10674--10685. {IEEE}, 2022.

\bibitem[Ronneberger et~al.(2015)Ronneberger, Fischer, and Brox]{DBLP:conf/miccai/RonnebergerFB15}
Olaf Ronneberger, Philipp Fischer, and Thomas Brox.
\newblock U-net: Convolutional networks for biomedical image segmentation.
\newblock In \emph{Medical Image Computing and Computer-Assisted Intervention - {MICCAI} 2015 - 18th International Conference Munich, Germany, October 5 - 9, 2015, Proceedings, Part {III}}, pages 234--241. Springer, 2015.

\bibitem[Roy et~al.(2020)Roy, Bhattacharya, Ghosh, and Pal]{DBLP:conf/cvpr/RoyBG020}
Prasun Roy, Saumik Bhattacharya, Subhankar Ghosh, and Umapada Pal.
\newblock {STEFANN:} scene text editor using font adaptive neural network.
\newblock In \emph{2020 {IEEE/CVF} Conference on Computer Vision and Pattern Recognition, {CVPR} 2020, Seattle, WA, USA, June 13-19, 2020}, pages 13225--13234. Computer Vision Foundation / {IEEE}, 2020.

\bibitem[Seitzer(2020)]{Seitzer2020FID}
Maximilian Seitzer.
\newblock {pytorch-fid: FID Score for PyTorch}.
\newblock \url{https://github.com/mseitzer/pytorch-fid}, 2020.
\newblock Version 0.3.0.

\bibitem[Sirignano and Spiliopoulos(2019)]{DBLP:journals/corr/abs-1907-04108}
Justin~A. Sirignano and Konstantinos Spiliopoulos.
\newblock Scaling limit of neural networks with the xavier initialization and convergence to a global minimum.
\newblock \emph{CoRR}, abs/1907.04108, 2019.

\bibitem[Song et~al.(2021)Song, Meng, and Ermon]{DBLP:conf/iclr/SongME21}
Jiaming Song, Chenlin Meng, and Stefano Ermon.
\newblock Denoising diffusion implicit models.
\newblock In \emph{9th International Conference on Learning Representations, {ICLR} 2021, Virtual Event, Austria, May 3-7, 2021}. OpenReview.net, 2021.

\bibitem[Su et~al.(2021)Su, Lu, Pan, Wen, and Liu]{DBLP:journals/corr/abs-2104-09864}
Jianlin Su, Yu Lu, Shengfeng Pan, Bo Wen, and Yunfeng Liu.
\newblock Roformer: Enhanced transformer with rotary position embedding.
\newblock \emph{CoRR}, abs/2104.09864, 2021.

\bibitem[Suvorov et~al.(2022)Suvorov, Logacheva, Mashikhin, Remizova, Ashukha, Silvestrov, Kong, Goka, Park, and Lempitsky]{DBLP:conf/wacv/SuvorovLMRASKGP22}
Roman Suvorov, Elizaveta Logacheva, Anton Mashikhin, Anastasia Remizova, Arsenii Ashukha, Aleksei Silvestrov, Naejin Kong, Harshith Goka, Kiwoong Park, and Victor Lempitsky.
\newblock Resolution-robust large mask inpainting with fourier convolutions.
\newblock In \emph{{IEEE/CVF} Winter Conference on Applications of Computer Vision, {WACV} 2022, Waikoloa, HI, USA, January 3-8, 2022}, pages 3172--3182. {IEEE}, 2022.

\bibitem[Tuo et~al.(2023)Tuo, Xiang, He, Geng, and Xie]{DBLP:journals/corr/abs-2311-03054}
Yuxiang Tuo, Wangmeng Xiang, Jun{-}Yan He, Yifeng Geng, and Xuansong Xie.
\newblock Anytext: Multilingual visual text generation and editing.
\newblock \emph{CoRR}, abs/2311.03054, 2023.

\bibitem[Tuo et~al.(2024)Tuo, Xiang, He, Geng, and Xie]{DBLP:conf/iclr/TuoXHGX24}
Yuxiang Tuo, Wangmeng Xiang, Jun{-}Yan He, Yifeng Geng, and Xuansong Xie.
\newblock Anytext: Multilingual visual text generation and editing.
\newblock In \emph{The Twelfth International Conference on Learning Representations, {ICLR} 2024, Vienna, Austria, May 7-11, 2024}. OpenReview.net, 2024.

\bibitem[Wu et~al.(2019)Wu, Zhang, Liu, Han, Liu, Ding, and Bai]{DBLP:conf/mm/WuZLHLDB19}
Liang Wu, Chengquan Zhang, Jiaming Liu, Junyu Han, Jingtuo Liu, Errui Ding, and Xiang Bai.
\newblock Editing text in the wild.
\newblock In \emph{Proceedings of the 27th {ACM} International Conference on Multimedia, {MM} 2019, Nice, France, October 21-25, 2019}, pages 1500--1508. {ACM}, 2019.

\bibitem[Yang et~al.(2020)Yang, Huang, and Lin]{DBLP:conf/cvpr/YangHL20}
Qiangpeng Yang, Jun Huang, and Wei Lin.
\newblock Swaptext: Image based texts transfer in scenes.
\newblock In \emph{2020 {IEEE/CVF} Conference on Computer Vision and Pattern Recognition, {CVPR} 2020, Seattle, WA, USA, June 13-19, 2020}, pages 14688--14697. Computer Vision Foundation / {IEEE}, 2020.

\bibitem[Yang et~al.(2023)Yang, Gui, Yuan, Liang, Ding, Hu, and Chen]{DBLP:conf/nips/YangGYLDH023}
Yukang Yang, Dongnan Gui, Yuhui Yuan, Weicong Liang, Haisong Ding, Han Hu, and Kai Chen.
\newblock Glyphcontrol: Glyph conditional control for visual text generation.
\newblock In \emph{Advances in Neural Information Processing Systems 36: Annual Conference on Neural Information Processing Systems 2023, NeurIPS 2023, New Orleans, LA, USA, December 10 - 16, 2023}, 2023.

\bibitem[Zeng et~al.(2024)Zeng, Shu, Li, Yang, and Zhou]{DBLP:conf/nips/ZengSLYZ24}
Weichao Zeng, Yan Shu, Zhenhang Li, Dongbao Yang, and Yu Zhou.
\newblock Textctrl: Diffusion-based scene text editing with prior guidance control.
\newblock In \emph{Advances in Neural Information Processing Systems 38: Annual Conference on Neural Information Processing Systems 2024, NeurIPS 2024, Vancouver, BC, Canada, December 10 - 15, 2024}, 2024.

\bibitem[Zhang et~al.(2023)Zhang, Rao, and Agrawala]{DBLP:conf/iccv/ZhangRA23}
Lvmin Zhang, Anyi Rao, and Maneesh Agrawala.
\newblock Adding conditional control to text-to-image diffusion models.
\newblock In \emph{{IEEE/CVF} International Conference on Computer Vision, {ICCV} 2023, Paris, France, October 1-6, 2023}, pages 3813--3824. {IEEE}, 2023.

\bibitem[Zhang et~al.(2018)Zhang, Isola, Efros, Shechtman, and Wang]{zhang2018perceptual}
Richard Zhang, Phillip Isola, Alexei~A Efros, Eli Shechtman, and Oliver Wang.
\newblock The unreasonable effectiveness of deep features as a perceptual metric.
\newblock In \emph{CVPR}, 2018.

\end{thebibliography}
}

\clearpage
\setcounter{page}{1}

\maketitlesupplementary

\section{Effects of Classifier-Free Guidance}

Classifier-free guidance (CFG) has demonstrated effectiveness in controlling the strength of prompt-following behavior in text-to-image diffusion models. Recognizing its potential utility in scene text editing, we incorporate CFG by training our model with a probabilistic null glyph condition (we also trained DiffUTE with CFG for fair comparison, though not in their original work).

\begin{figure}[h]
\centering
\begin{tabular}{@{}c@{\hspace{1mm}}c@{}}
\includegraphics[width=0.22\textwidth]{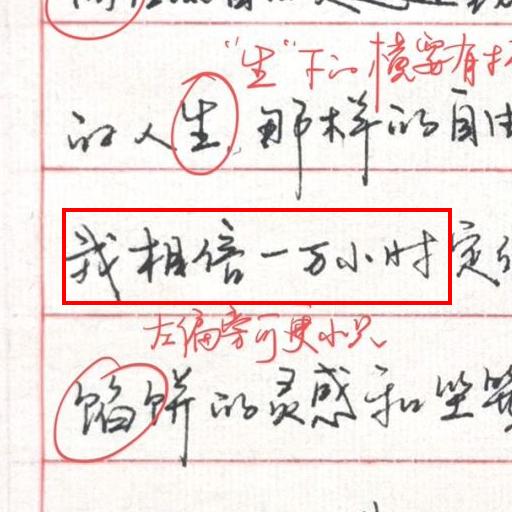} &
\includegraphics[width=0.22\textwidth]{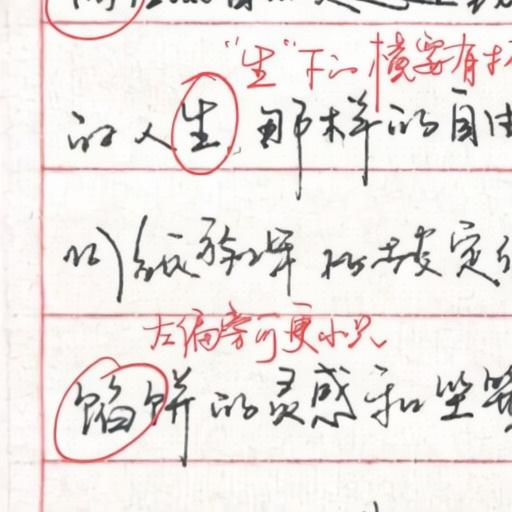} \\
original & CFG=1 \\[6pt]
\includegraphics[width=0.22\textwidth]{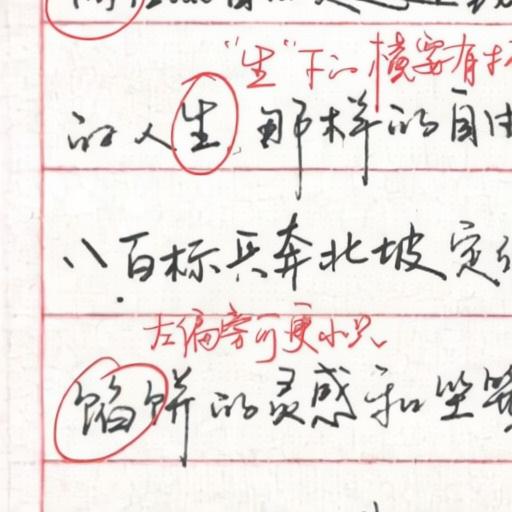} &
\includegraphics[width=0.22\textwidth]{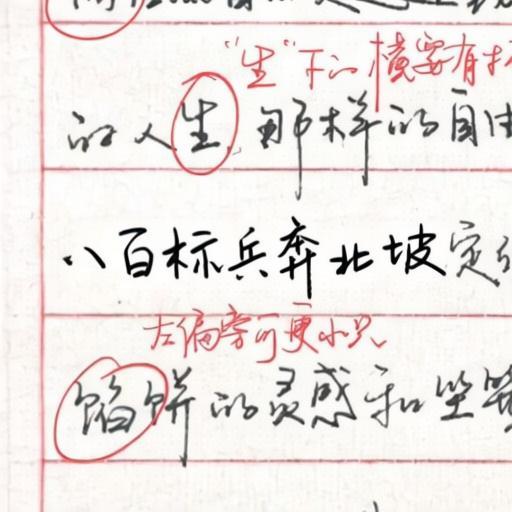} \\
CFG=3 & CFG=5 \\[6pt]
\end{tabular}
Prompt: "\includegraphics{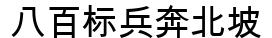}" 
\caption{Effect of classifier-free guidance (CFG). Original image with target area mask shown top-left. Without CFG (i.e. CFG=1), GlyphMastero produces unreadable text. CFG=3 improves readability while maintaining style. CFG=5 generates overly thick text, deviating from the original region.}
\label{fig:cfg}
\end{figure}

Our experiments with CFG reveal a crucial trade-off in scene text editing. As demonstrated in Figure \ref{fig:cfg}, we found that in inference, a higher CFG scale results in stronger glyph control, producing clearer and thicker text. This allows for improved readability when editing texts. However, our findings show that this comes at a cost to style preservation. Conversely, lower CFG scales excel at maintaining the original text style, though occasionally at the expense of target text accuracy. This insight offers a new approach to balancing readability and style preservation in the scene text editing task.

\section{Example Failure Cases}
\begin{figure}[h]
\centering
\includegraphics[width=0.3\textwidth]{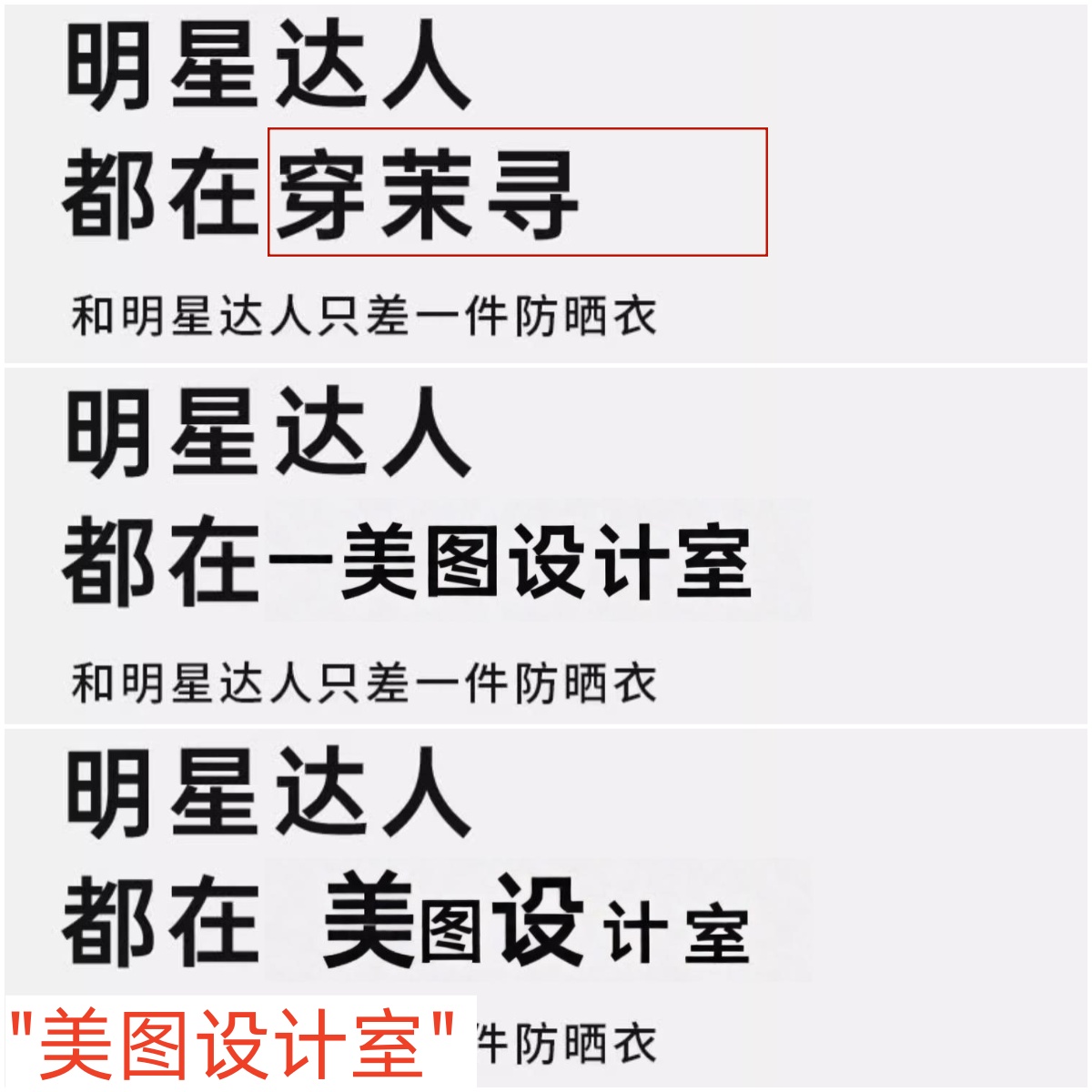} 
\caption{Example of a failure case. The upper image displays the source text with regions marked by red boxes. The middle and bottom images show two unsuccessful generation attempts. The intended target text appears at the bottom left of each generated result.}
\label{fig:failure}
\end{figure}

As shown in Figure \ref{fig:failure}, our method encounters limitations when the selected editing region substantially exceeds the target text length. In such scenarios, the model struggles to maintain coherent text generation, resulting in irregularly sized characters and occasional repetition patterns in the output. These artifacts emerge as the model attempts to distribute textual elements across disproportionately large spatial regions.

\section{Additional Results}

\subsection{Quantitative Comparison}
To cross-validate the effectiveness of our method, we also evaluated our method on TextCtrl's ScenePair dataset (1,285 test cases), re-evaluating all other methods using TextCtrl's GitHub-published results and scripts. Table~\ref{tab:sp_combined} showed our superior generation accuracy and strong performance across style metrics, except for a slightly higher FID.

\begin{table}[h]
    \centering
    \setlength{\tabcolsep}{3pt}
    \footnotesize
    \begin{tabular}{l|cc|cccc}
    \hline
    \multirow{2}{*}{Methods} & \multicolumn{2}{c|}{Accuracy} & \multicolumn{4}{c}{Similarity} \\
    \cline{2-7}
     & W.Acc$\uparrow$ & NED$\uparrow$ & SSIM$\uparrow$ & PSNR$\uparrow$ & MSE$\downarrow$ & FID$\downarrow$ \\
    \hline
    SRNet & 16.64 & 0.4790 & 26.66 & 14.08 & 5.61 & 49.23 \\
    MOSTEL & 35.16 & 0.5570 & 27.46 & 14.46 & 5.19 & 49.20 \\
    DiffSTE & 29.14 & 0.5255 & 26.91 & 13.49 & 6.07 & 118.60 \\
    TextDiffuser & 51.48 & 0.7190 & 27.02 & 13.99 & 5.72 & 57.48 \\
    AnyText & 47.97 & 0.7186 & 31.19 & 13.58 & 6.36 & 52.07 \\
    TextCtrl & 78.91 & 0.9199 & 37.93 & 14.92 & 4.58 & \textbf{31.98} \\
    Ours & \textbf{83.52} & \textbf{0.9572} & \textbf{47.58} & \textbf{16.25} & \textbf{3.97} & 32.03 \\
    \hline
    \end{tabular}
    \caption{Performance comparison on English \emph{ScenePair} testset. SSIM and MSE scaled by $\times10^{-2}$. W.Acc: Word Accuracy.}
    \label{tab:sp_combined}
    \end{table}

\subsection{Qualitative Comparison}
\begin{figure*}[h]
\centering
\begin{tabular}{ccccc}
\emph{Prompt} & \emph{Masked Source Image} & {DiffUTE} & {AnyText} & \textbf{Ours} \\
\midrule
\raisebox{-.5\height}{\includegraphics{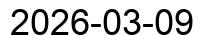}} &
\raisebox{-.5\height}{\includegraphics[width=0.16\textwidth]{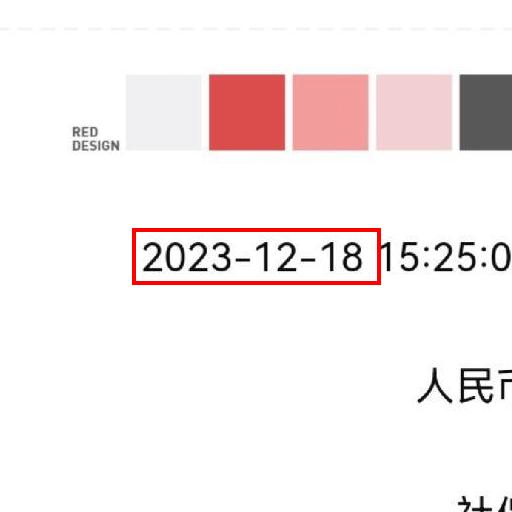}} &
\raisebox{-.5\height}{\includegraphics[width=0.16\textwidth]{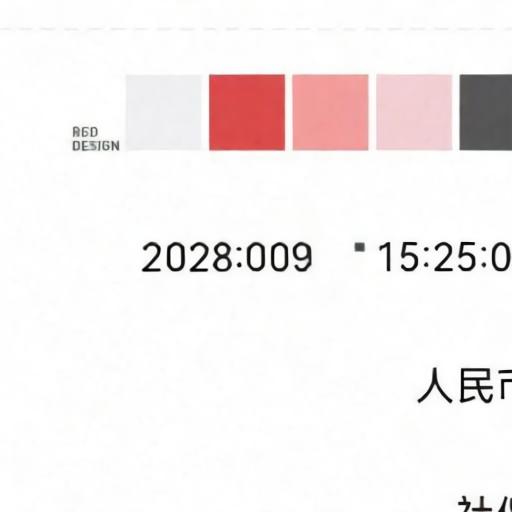}} &
\raisebox{-.5\height}{\includegraphics[width=0.16\textwidth]{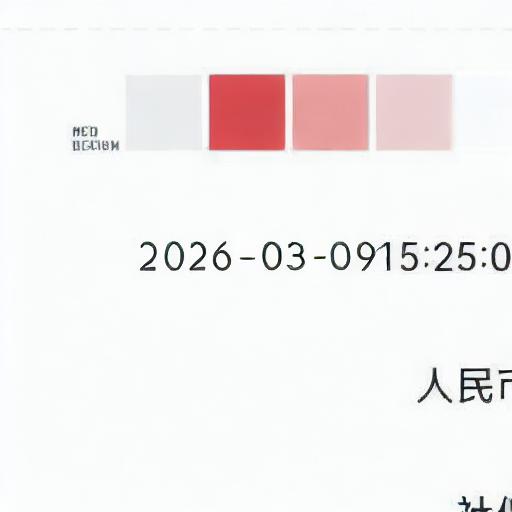}} &
\raisebox{-.5\height}{\includegraphics[width=0.16\textwidth]{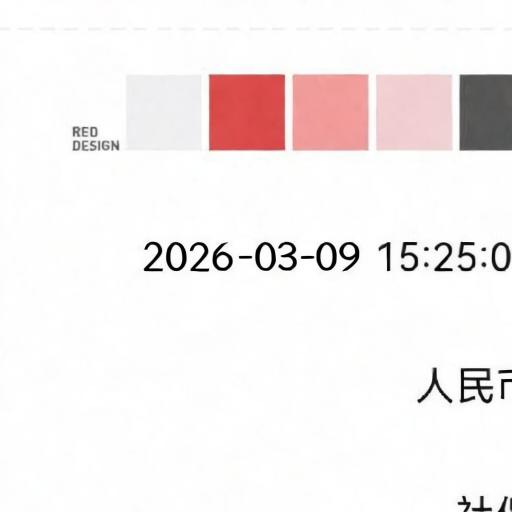}} \\
\raisebox{-.5\height}{
\shortstack{
\includegraphics{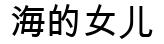} \\ \textit{(the daughter of the sea)}}
} &
\raisebox{-.5\height}{\includegraphics[width=0.16\textwidth]{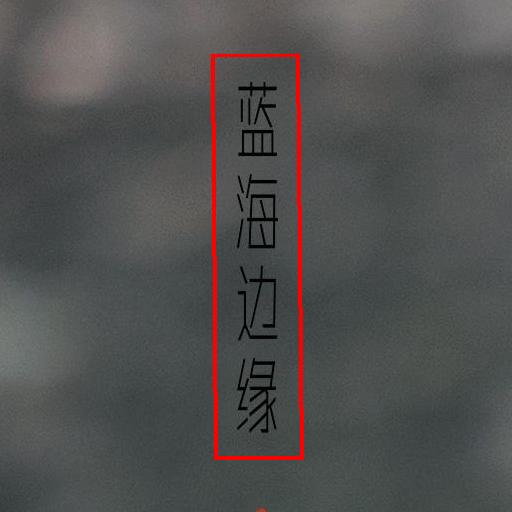}} &
\raisebox{-.5\height}{\includegraphics[width=0.16\textwidth]{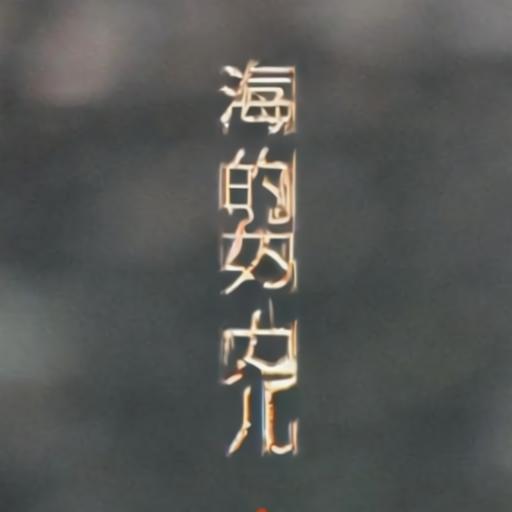}} &
\raisebox{-.5\height}{\includegraphics[width=0.16\textwidth]{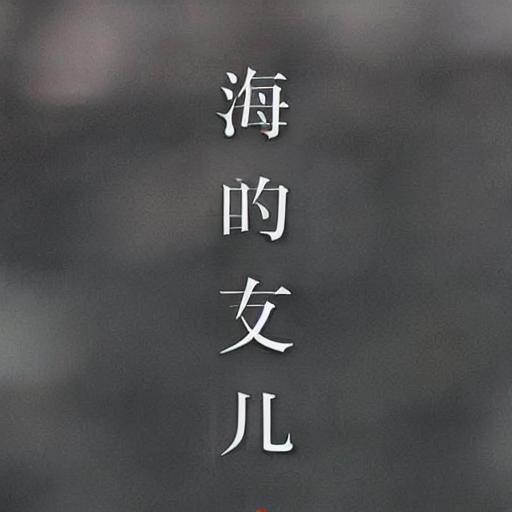}} &
\raisebox{-.5\height}{\includegraphics[width=0.16\textwidth]{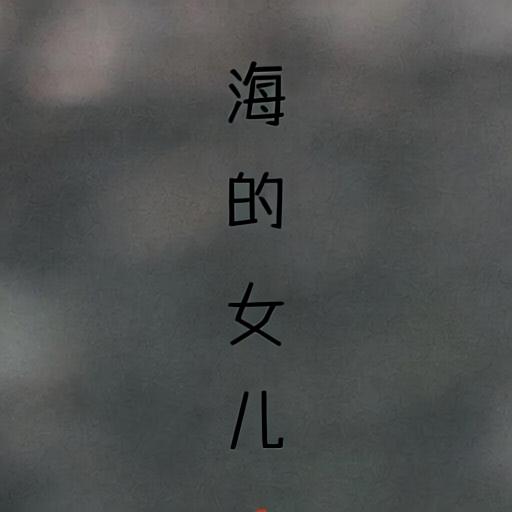}} \\
\raisebox{-.5\height}{\includegraphics{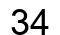}} &
\raisebox{-.5\height}{\includegraphics[width=0.16\textwidth]{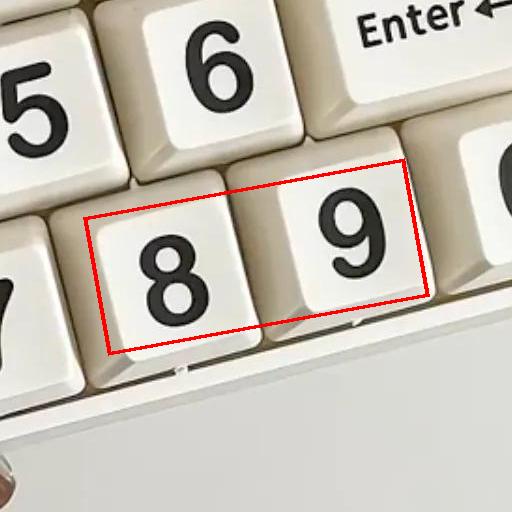}} &
\raisebox{-.5\height}{\includegraphics[width=0.16\textwidth]{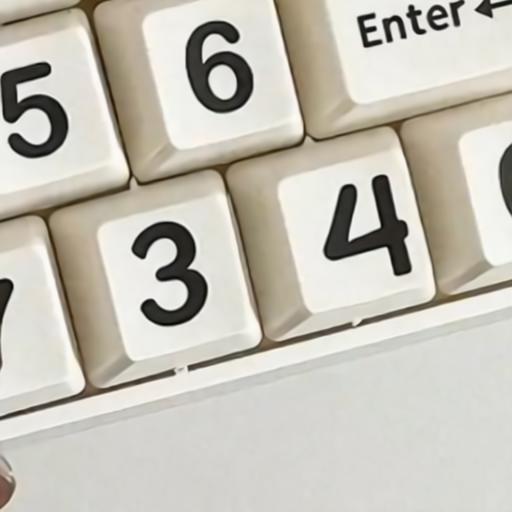}} &
\raisebox{-.5\height}{\includegraphics[width=0.16\textwidth]{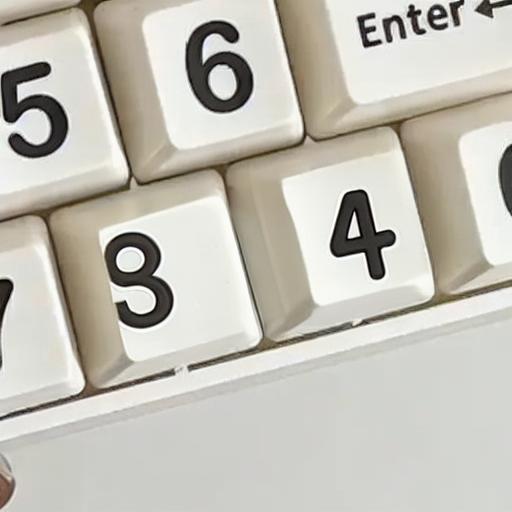}} &
\raisebox{-.5\height}{\includegraphics[width=0.16\textwidth]{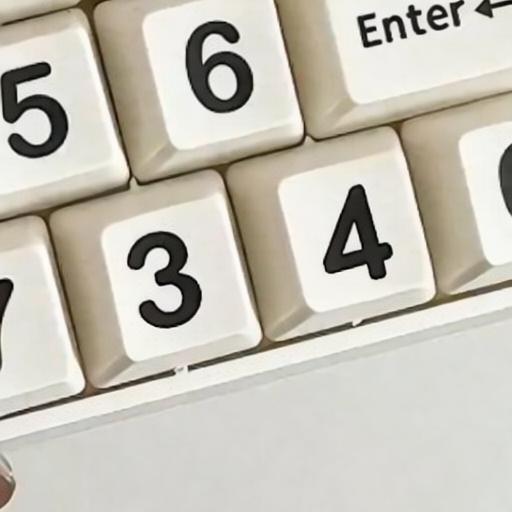}} \\
\raisebox{-.5\height}{
\shortstack{
\includegraphics{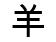} \\
\textit{(sheep)}}
} &
\raisebox{-.5\height}{\includegraphics[width=0.16\textwidth]{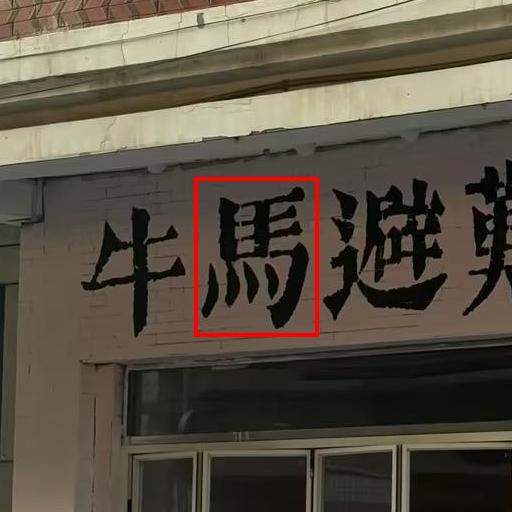}} &
\raisebox{-.5\height}{\includegraphics[width=0.16\textwidth]{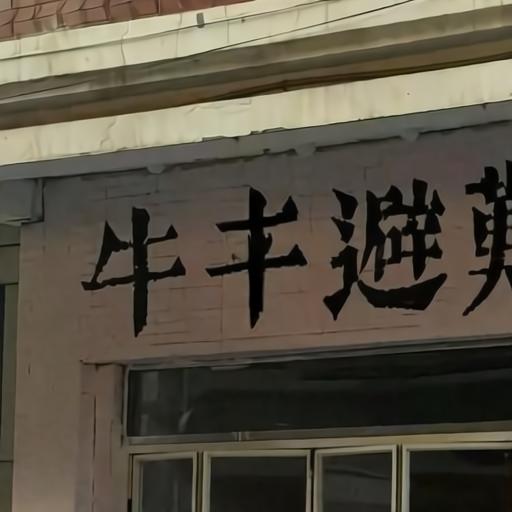}} &
\raisebox{-.5\height}{\includegraphics[width=0.16\textwidth]{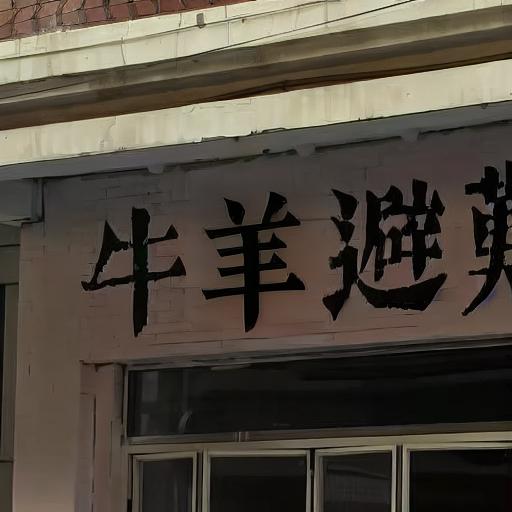}} &
\raisebox{-.5\height}{\includegraphics[width=0.16\textwidth]{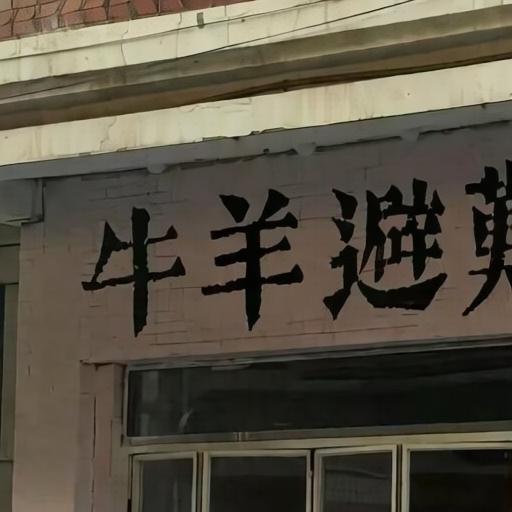}} \\
\raisebox{-.5\height}{
\shortstack{
\includegraphics{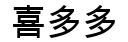} \\
\textit{(Xi'duo'duo)}}
} &
\raisebox{-.5\height}{\includegraphics[width=0.16\textwidth]{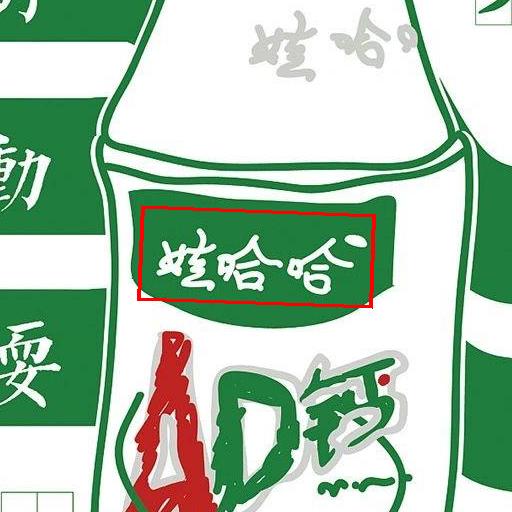}} &
\raisebox{-.5\height}{\includegraphics[width=0.16\textwidth]{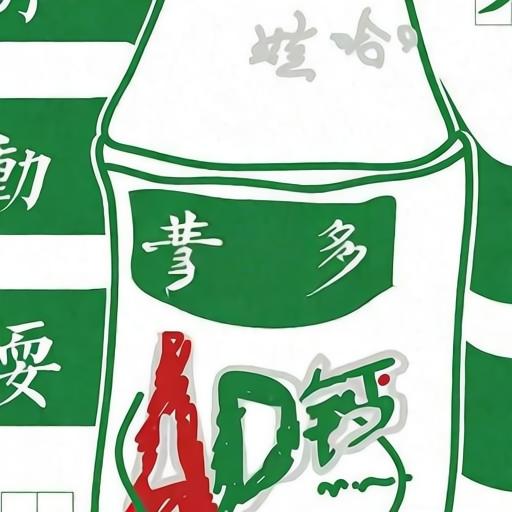}} &
\raisebox{-.5\height}{\includegraphics[width=0.16\textwidth]{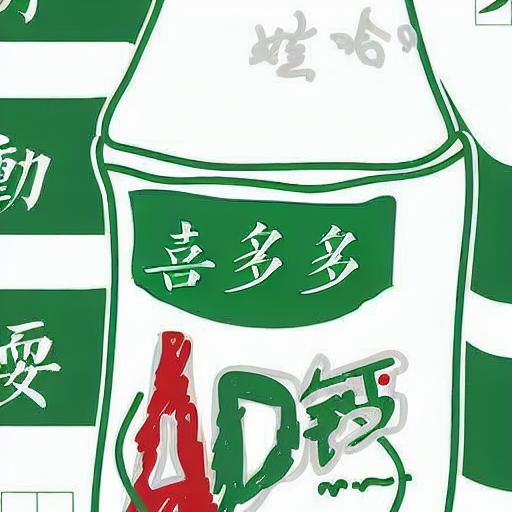}} &
\raisebox{-.5\height}{\includegraphics[width=0.16\textwidth]{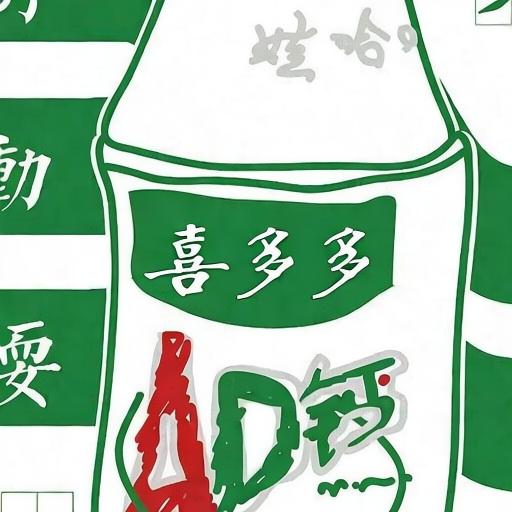}} \\
\raisebox{-.5\height}{
\shortstack{
\includegraphics{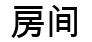} \\
\textit{(room)}}
} &
\raisebox{-.5\height}{\includegraphics[width=0.16\textwidth]{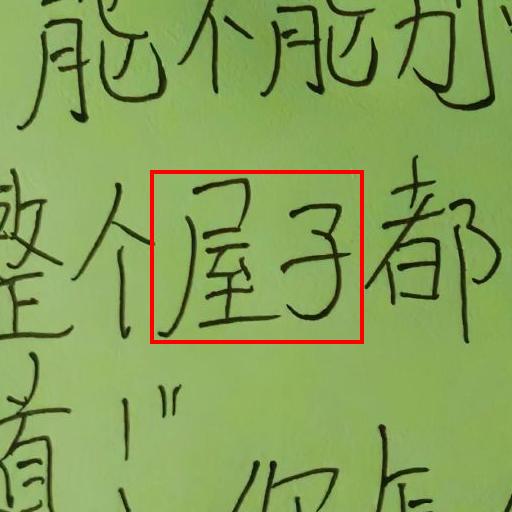}} &
\raisebox{-.5\height}{\includegraphics[width=0.16\textwidth]{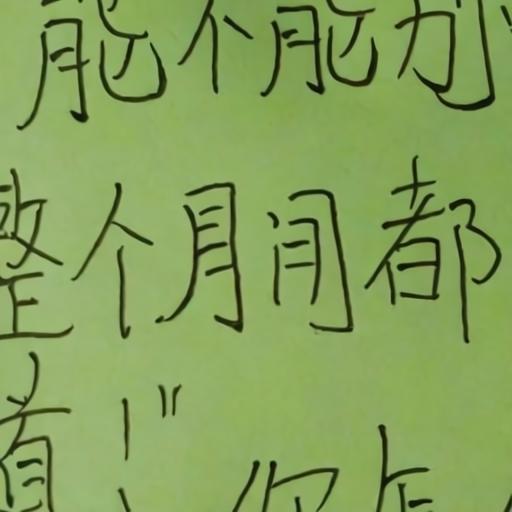}} &
\raisebox{-.5\height}{\includegraphics[width=0.16\textwidth]{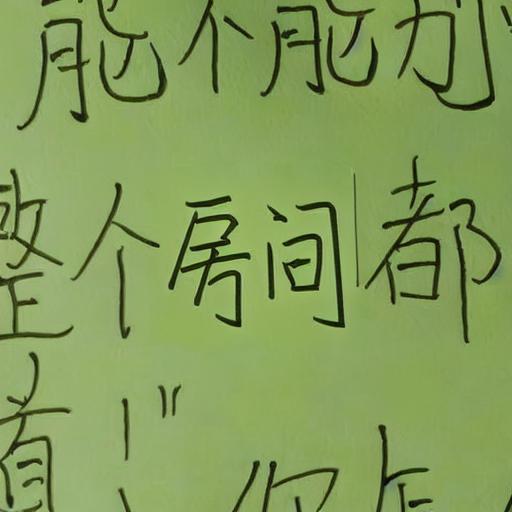}} &
\raisebox{-.5\height}{\includegraphics[width=0.16\textwidth]{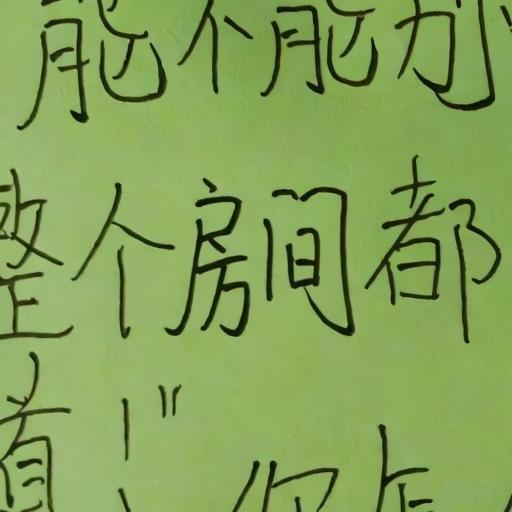}} \\
\raisebox{-.5\height}{
\shortstack{
\includegraphics{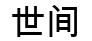} \\
\textit{(the world)}}
} &
\raisebox{-.5\height}{\includegraphics[width=0.16\textwidth]{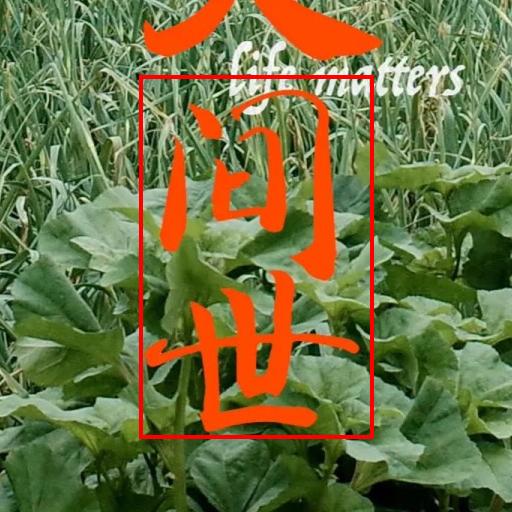}} &
\raisebox{-.5\height}{\includegraphics[width=0.16\textwidth]{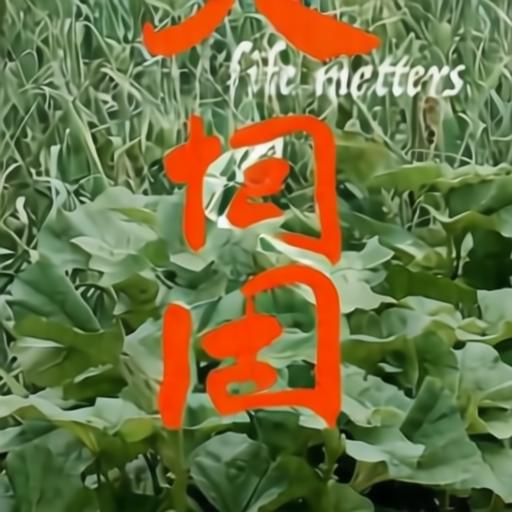}} &
\raisebox{-.5\height}{\includegraphics[width=0.16\textwidth]{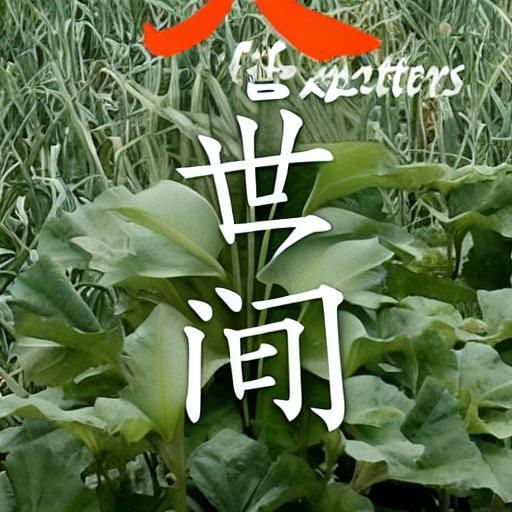}} &
\raisebox{-.5\height}{\includegraphics[width=0.16\textwidth]{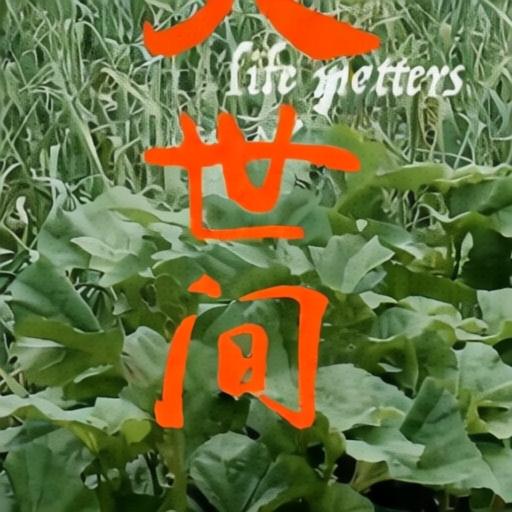}} \\
\end{tabular}
\caption{Comparison results on our test set with stylish scene texts}
\label{fig:ours}
\end{figure*}

\begin{figure*}[h]
\centering
\begin{tabular}{ccccc}
\emph{Prompt} & \emph{Masked Source Image} & {DiffUTE} & {AnyText} & \textbf{Ours} \\
\midrule
\raisebox{-.5\height}{\includegraphics{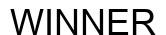}} &
\raisebox{-.5\height}{\includegraphics[width=0.16\textwidth]{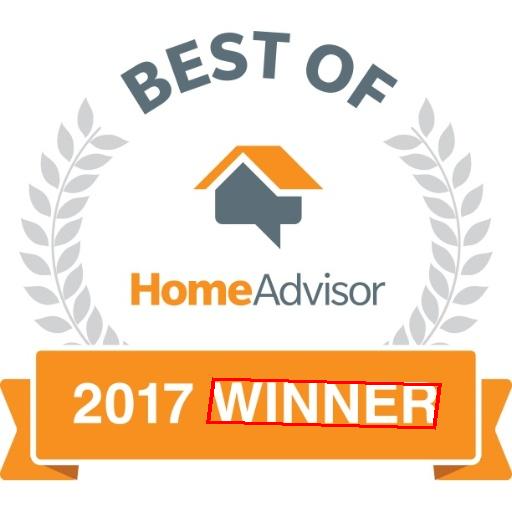}} &
\raisebox{-.5\height}{\includegraphics[width=0.16\textwidth]{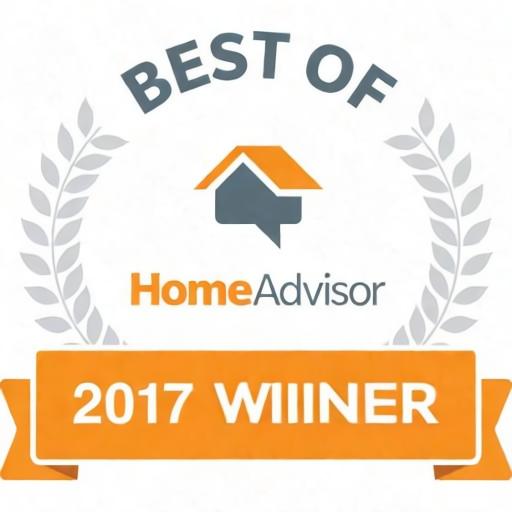}} &
\raisebox{-.5\height}{\includegraphics[width=0.16\textwidth]{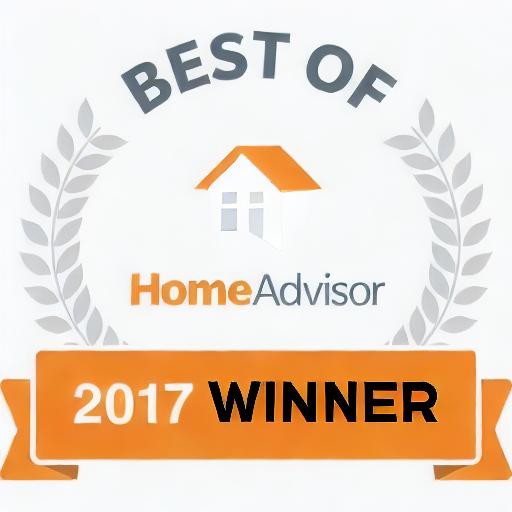}} &
\raisebox{-.5\height}{\includegraphics[width=0.16\textwidth]{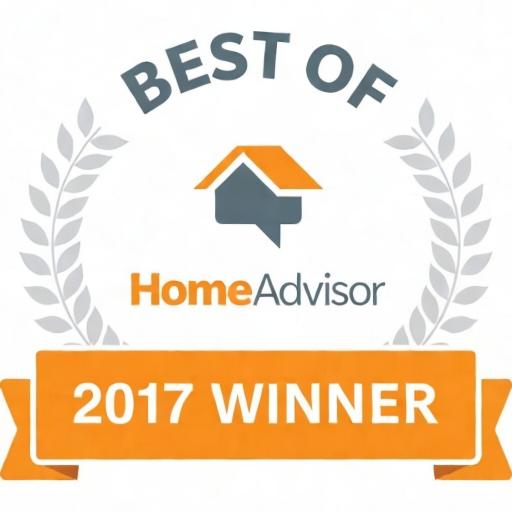}} \\
\raisebox{-.5\height}{\includegraphics{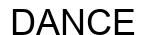}} &
\raisebox{-.5\height}{\includegraphics[width=0.16\textwidth]{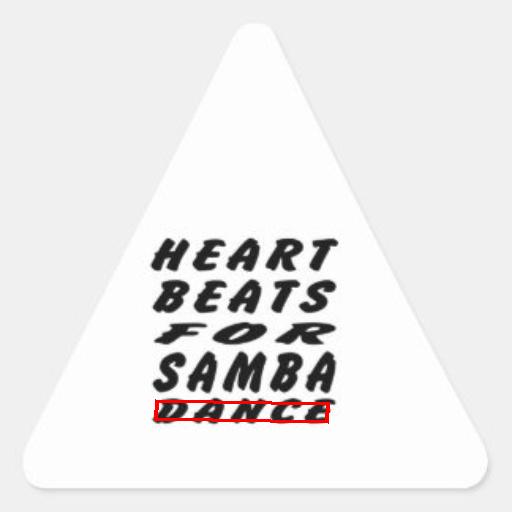}} &
\raisebox{-.5\height}{\includegraphics[width=0.16\textwidth]{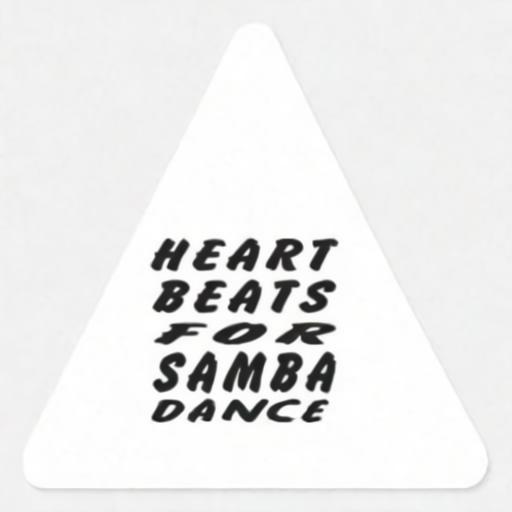}} &
\raisebox{-.5\height}{\includegraphics[width=0.16\textwidth]{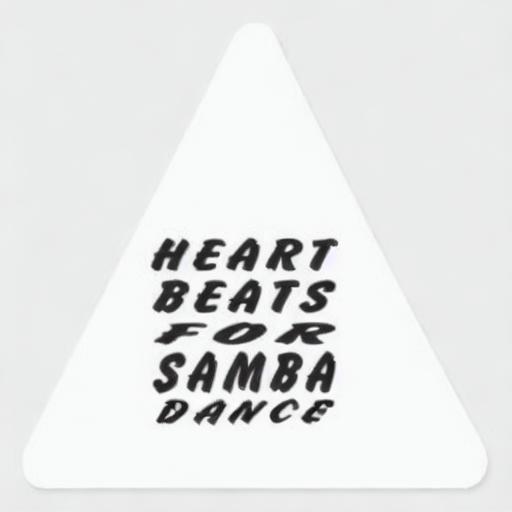}} &
\raisebox{-.5\height}{\includegraphics[width=0.16\textwidth]{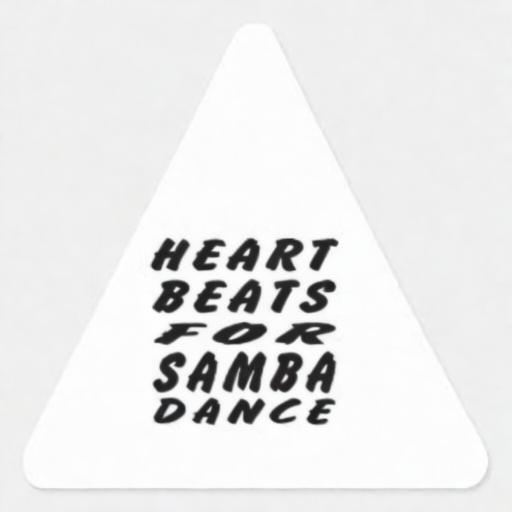}} \\
\raisebox{-.5\height}
{\includegraphics{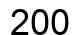}} &
\raisebox{-.5\height}{\includegraphics[width=0.16\textwidth]{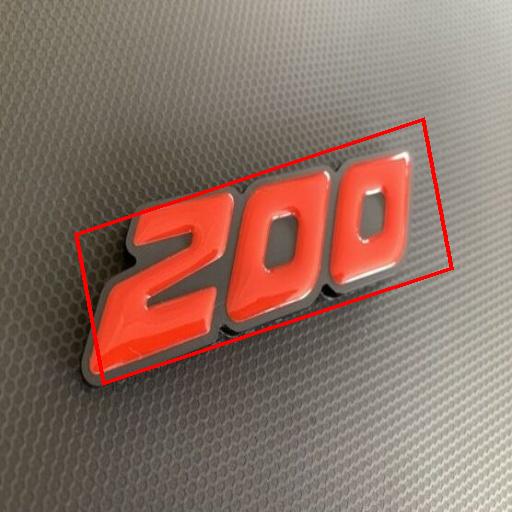}} &
\raisebox{-.5\height}{\includegraphics[width=0.16\textwidth]{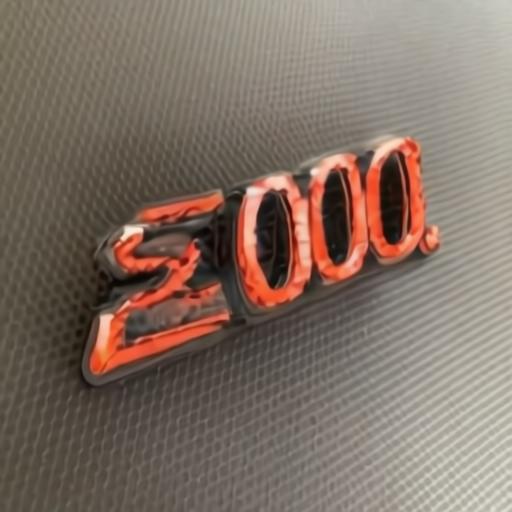}} &
\raisebox{-.5\height}{\includegraphics[width=0.16\textwidth]{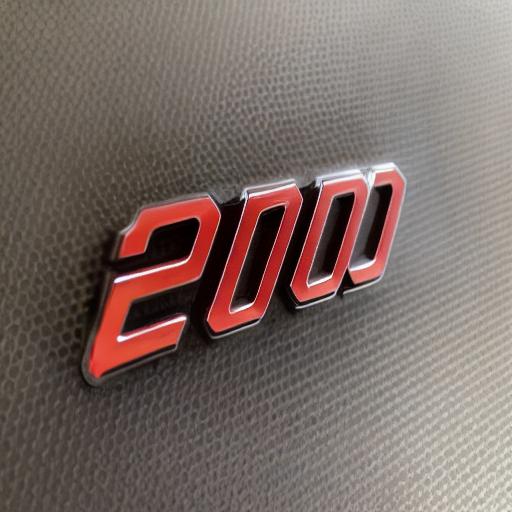}} &
\raisebox{-.5\height}{\includegraphics[width=0.16\textwidth]{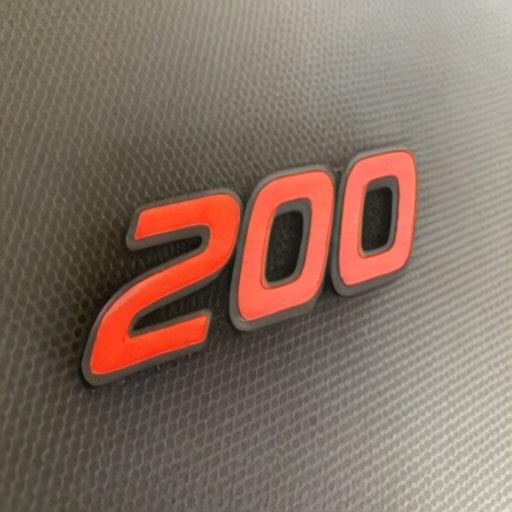}} \\
\raisebox{-.5\height}{\includegraphics{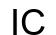}} &
\raisebox{-.5\height}{\includegraphics[width=0.16\textwidth]{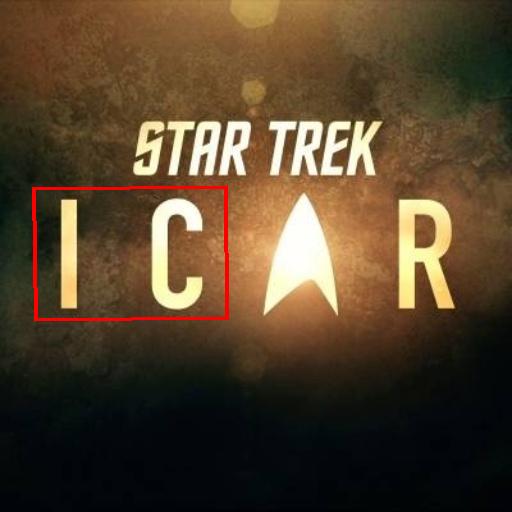}} &
\raisebox{-.5\height}{\includegraphics[width=0.16\textwidth]{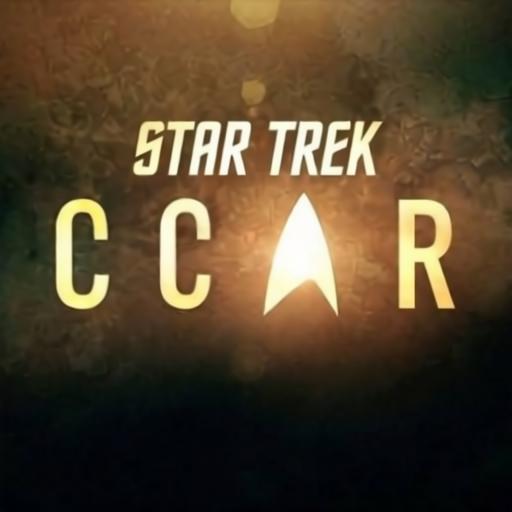}} &
\raisebox{-.5\height}{\includegraphics[width=0.16\textwidth]{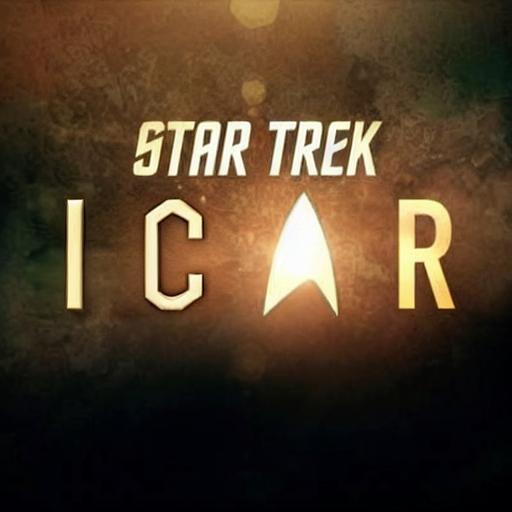}} &
\raisebox{-.5\height}{\includegraphics[width=0.16\textwidth]{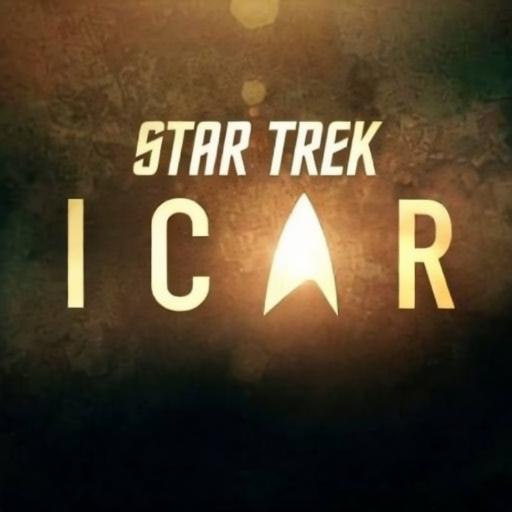}} \\
\raisebox{-.5\height}{\includegraphics{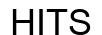}} &
\raisebox{-.5\height}{\includegraphics[width=0.16\textwidth]{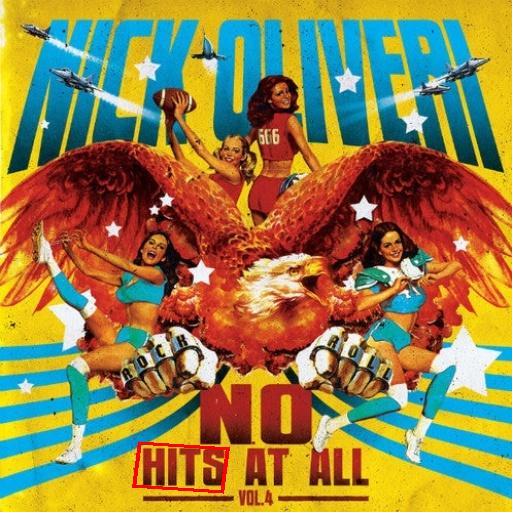}} &
\raisebox{-.5\height}{\includegraphics[width=0.16\textwidth]{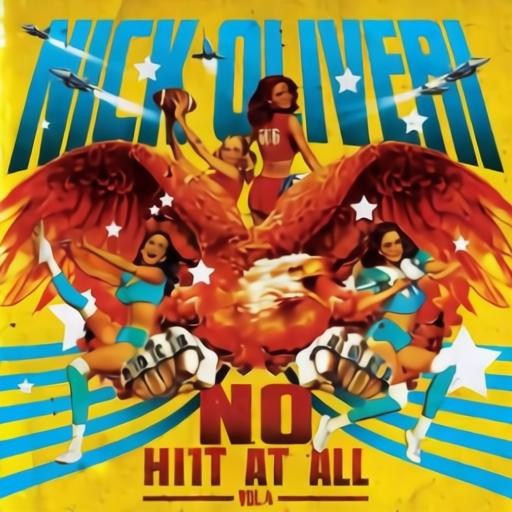}} &
\raisebox{-.5\height}{\includegraphics[width=0.16\textwidth]{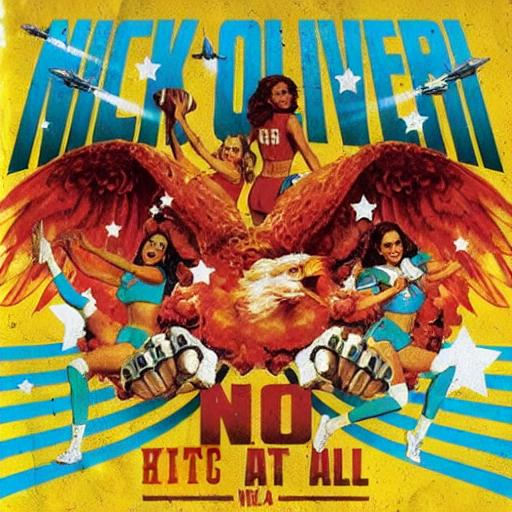}} &
\raisebox{-.5\height}{\includegraphics[width=0.16\textwidth]{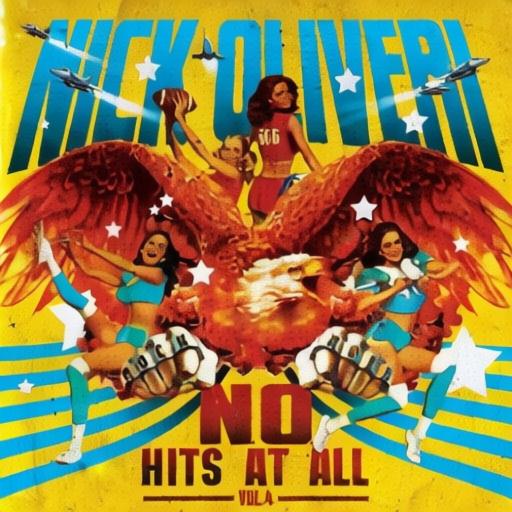}} \\
\raisebox{-.5\height}{\includegraphics{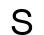}} &
\raisebox{-.5\height}{\includegraphics[width=0.16\textwidth]{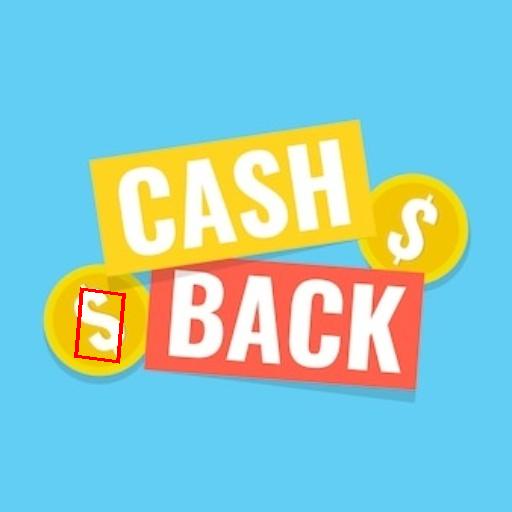}} &
\raisebox{-.5\height}{\includegraphics[width=0.16\textwidth]{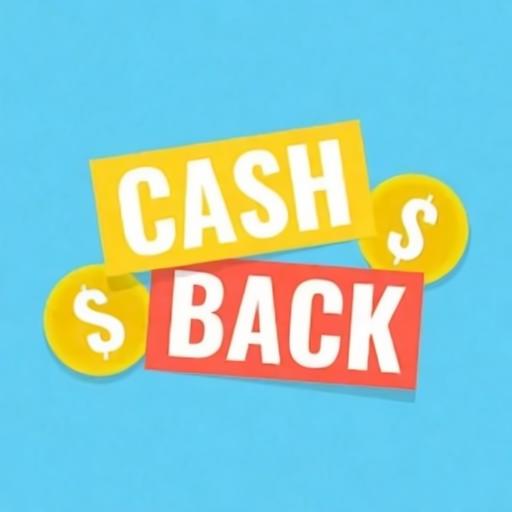}} &
\raisebox{-.5\height}{\includegraphics[width=0.16\textwidth]{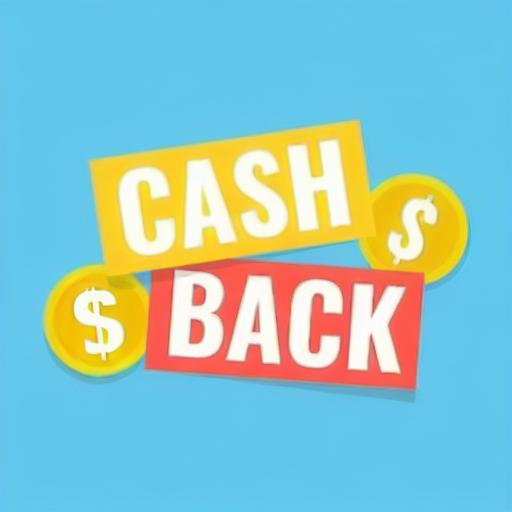}} &
\raisebox{-.5\height}{\includegraphics[width=0.16\textwidth]{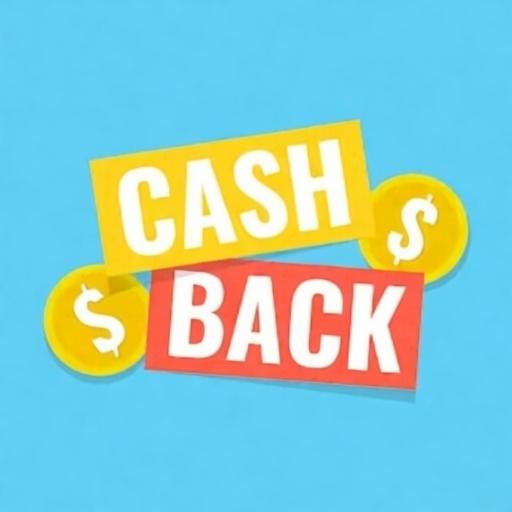}} \\
\raisebox{-.5\height}{\includegraphics{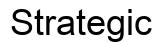}} &
\raisebox{-.5\height}{\includegraphics[width=0.16\textwidth]{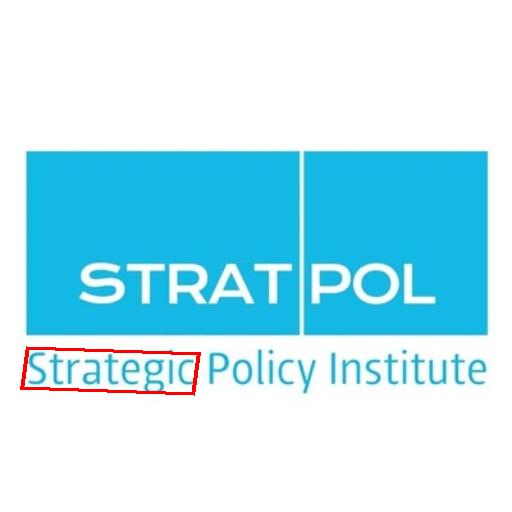}} &
\raisebox{-.5\height}{\includegraphics[width=0.16\textwidth]{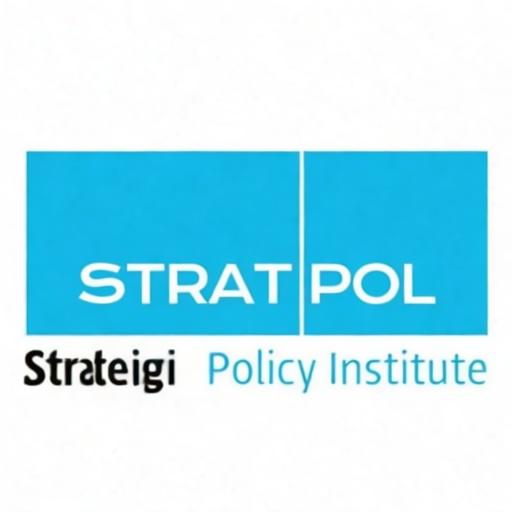}} &
\raisebox{-.5\height}{\includegraphics[width=0.16\textwidth]{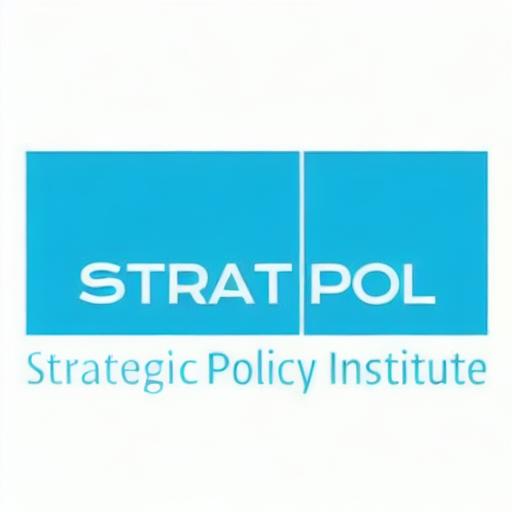}} &
\raisebox{-.5\height}{\includegraphics[width=0.16\textwidth]{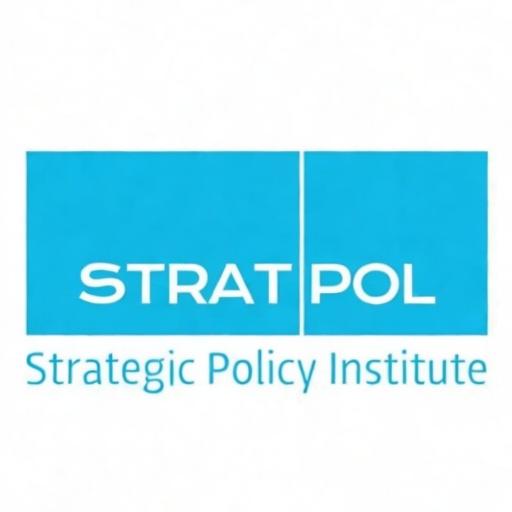}} \\
\end{tabular}
\caption{Comparison results on English (LAION) test set}
\label{fig:laion}
\end{figure*}

\begin{figure*}[h]
\centering
\begin{tabular}{ccccc}
\emph{Prompt} & \emph{Masked Source Image} & {DiffUTE} & {AnyText} & \textbf{Ours} \\
\midrule
\raisebox{-.5\height}{
\shortstack{
\includegraphics{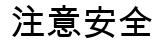} \\
\textit{(watch out)}}
} &
\raisebox{-.5\height}{\includegraphics[width=0.16\textwidth]{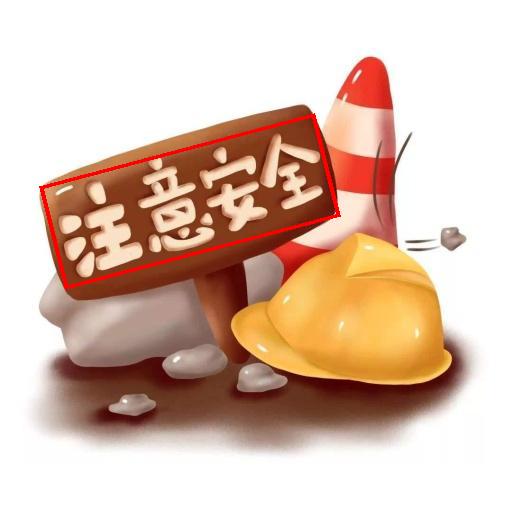}} &
\raisebox{-.5\height}{\includegraphics[width=0.16\textwidth]{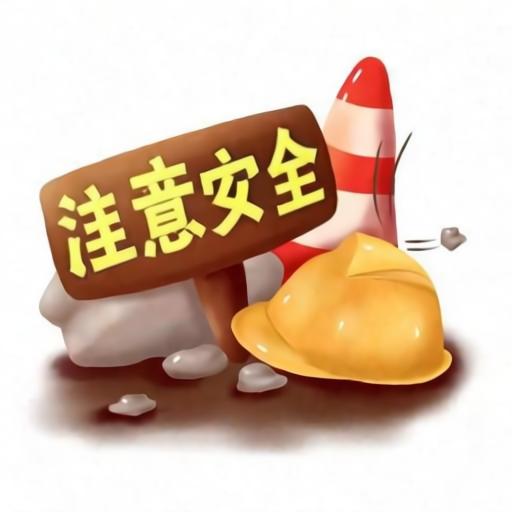}} &
\raisebox{-.5\height}{\includegraphics[width=0.16\textwidth]{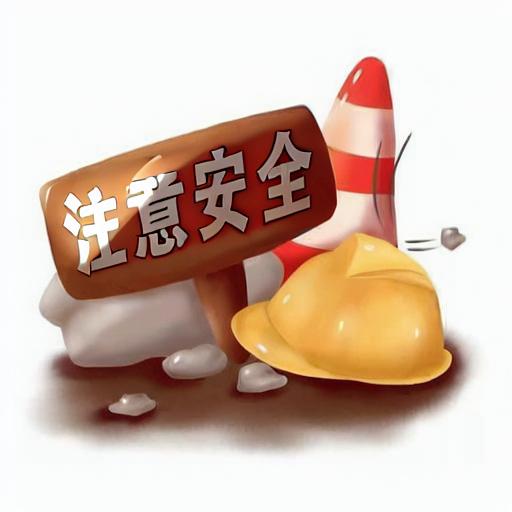}} &
\raisebox{-.5\height}{\includegraphics[width=0.16\textwidth]{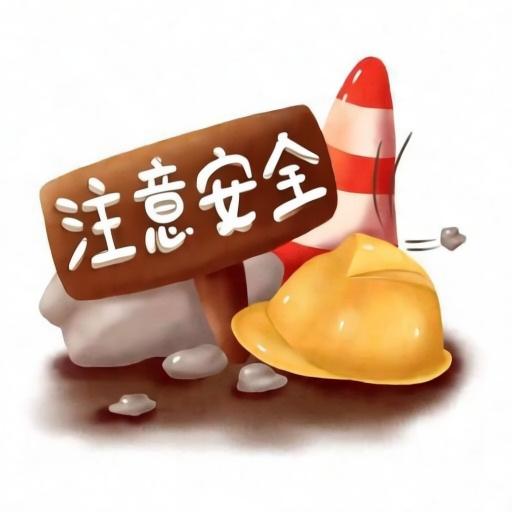}} \\
\raisebox{-.5\height}{
\shortstack{
\includegraphics{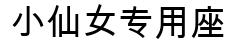} \\
\textit{(dedicated seat for girls)}}
} &
\raisebox{-.5\height}{\includegraphics[width=0.16\textwidth]{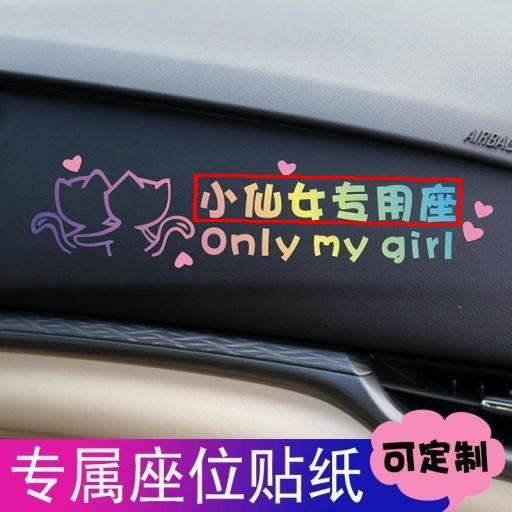}} &
\raisebox{-.5\height}{\includegraphics[width=0.16\textwidth]{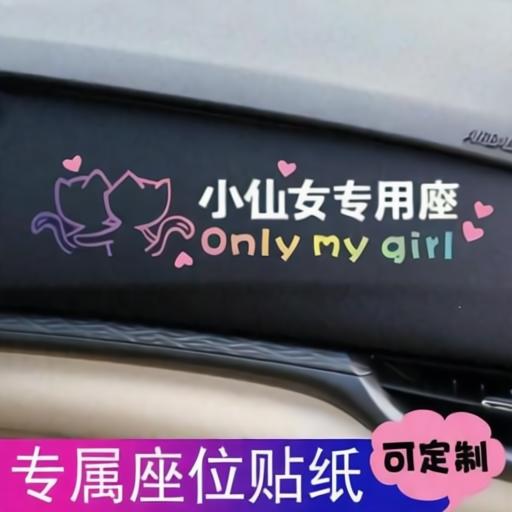}} &
\raisebox{-.5\height}{\includegraphics[width=0.16\textwidth]{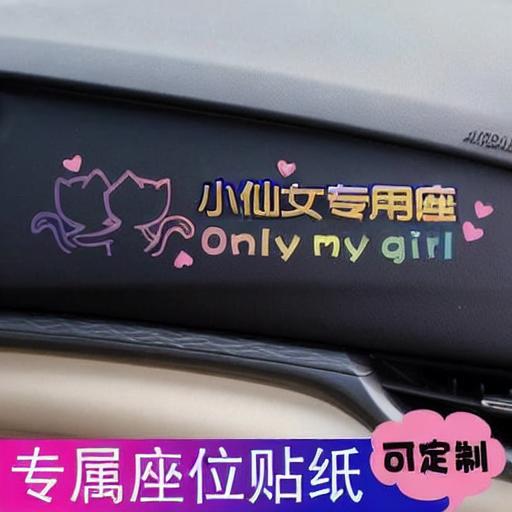}} &
\raisebox{-.5\height}{\includegraphics[width=0.16\textwidth]{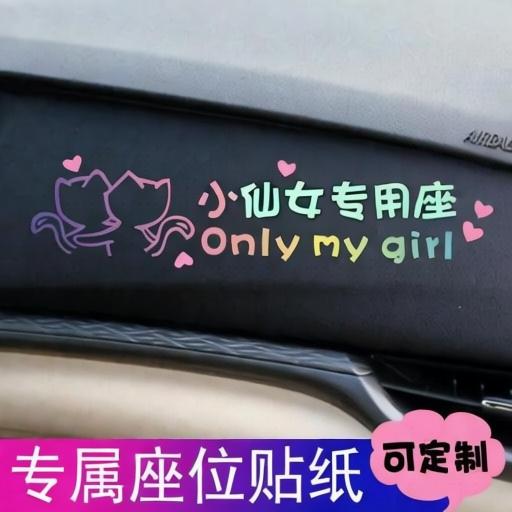}} \\
\raisebox{-.5\height}{
\shortstack{
\includegraphics{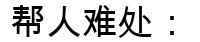} \\
\textit{(help others)}}
} &
\raisebox{-.5\height}{\includegraphics[width=0.16\textwidth]{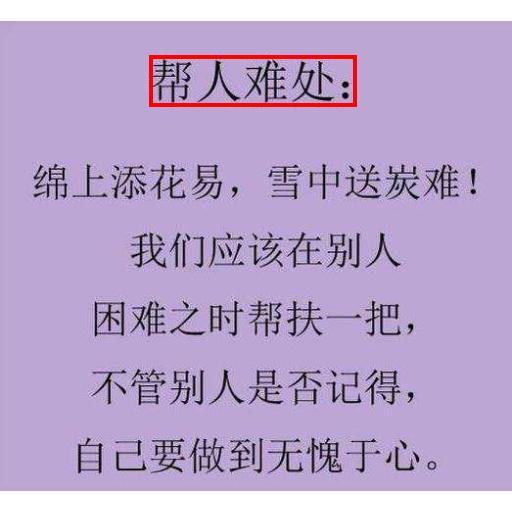}} &
\raisebox{-.5\height}{\includegraphics[width=0.16\textwidth]{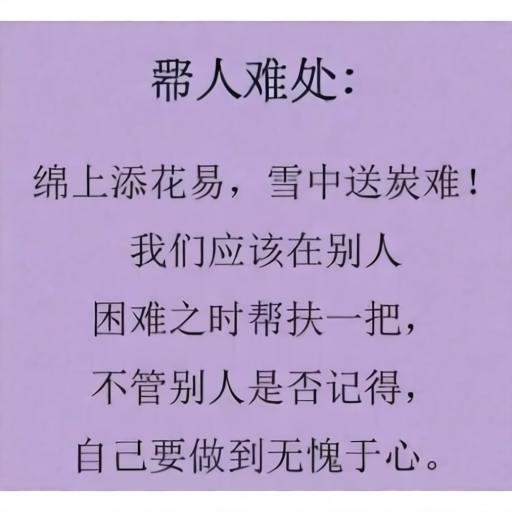}} &
\raisebox{-.5\height}{\includegraphics[width=0.16\textwidth]{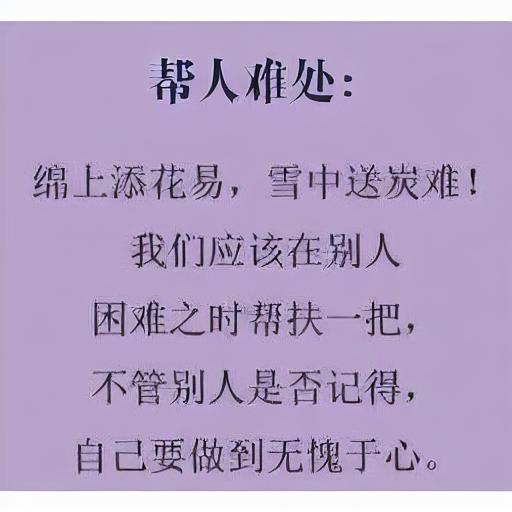}} &
\raisebox{-.5\height}{\includegraphics[width=0.16\textwidth]{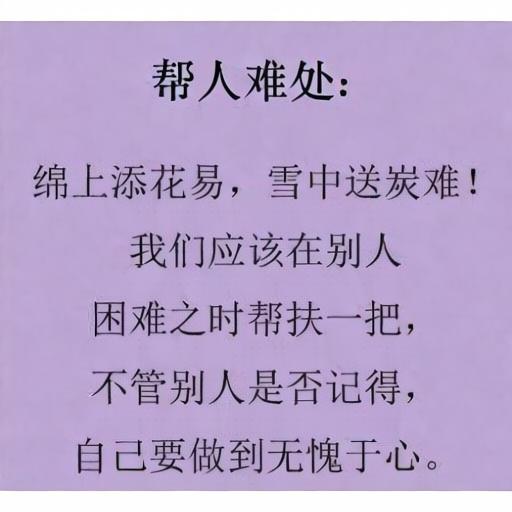}} \\
\raisebox{-.5\height}{
\shortstack{
\includegraphics{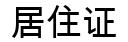} \\
\textit{(residency ID)}}
} &
\raisebox{-.5\height}{\includegraphics[width=0.16\textwidth]{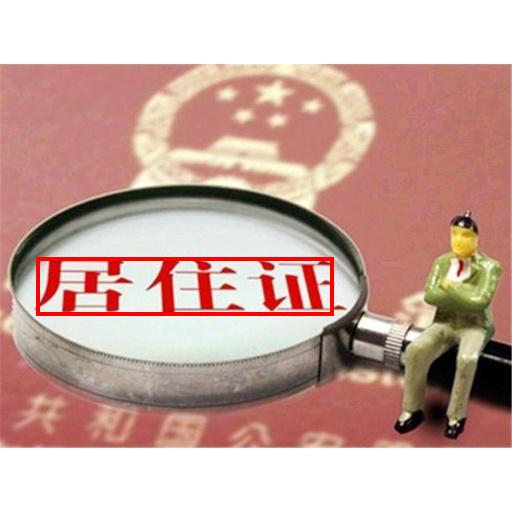}} &
\raisebox{-.5\height}{\includegraphics[width=0.16\textwidth]{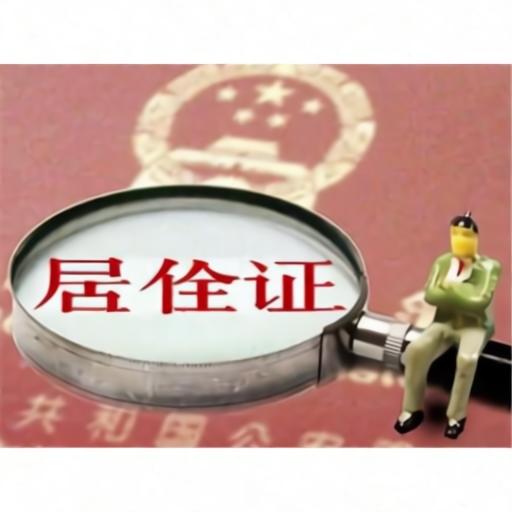}} &
\raisebox{-.5\height}{\includegraphics[width=0.16\textwidth]{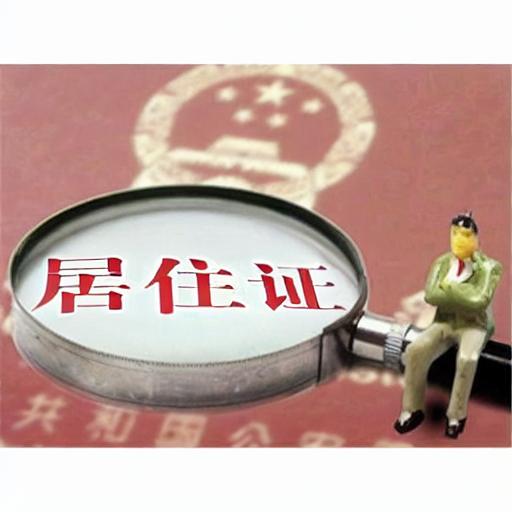}} &
\raisebox{-.5\height}{\includegraphics[width=0.16\textwidth]{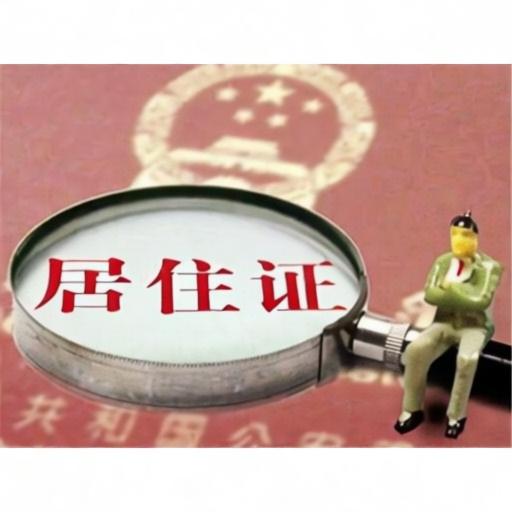}} \\
\raisebox{-.5\height}{
\shortstack{
\includegraphics{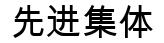} \\
\textit{(advanced group of people)}}
} &
\raisebox{-.5\height}{\includegraphics[width=0.16\textwidth]{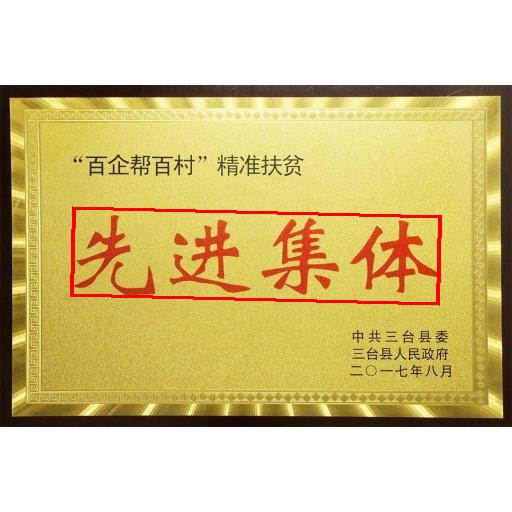}} &
\raisebox{-.5\height}{\includegraphics[width=0.16\textwidth]{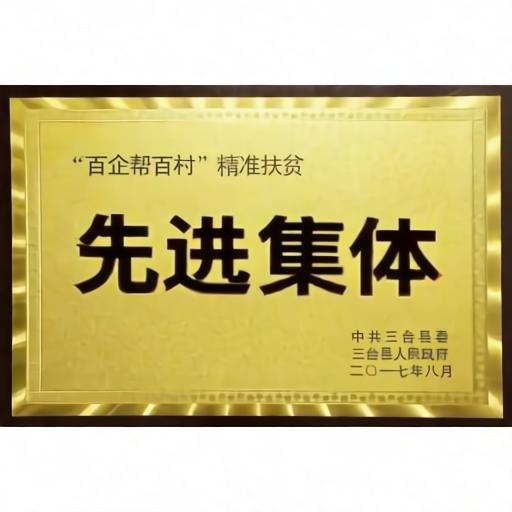}} &
\raisebox{-.5\height}{\includegraphics[width=0.16\textwidth]{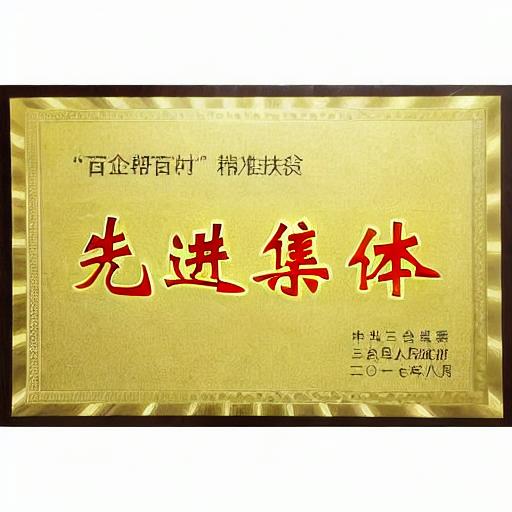}} &
\raisebox{-.5\height}{\includegraphics[width=0.16\textwidth]{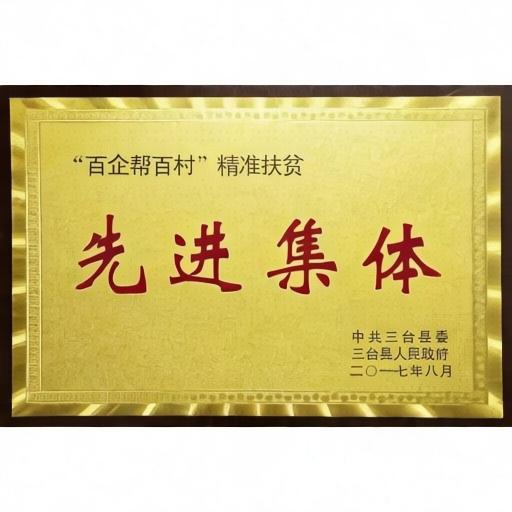}} \\
\raisebox{-.5\height}{
\shortstack{
\includegraphics{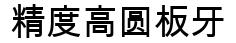} \\
\textit{(high precision bolt)}}
} &
\raisebox{-.5\height}{\includegraphics[width=0.16\textwidth]{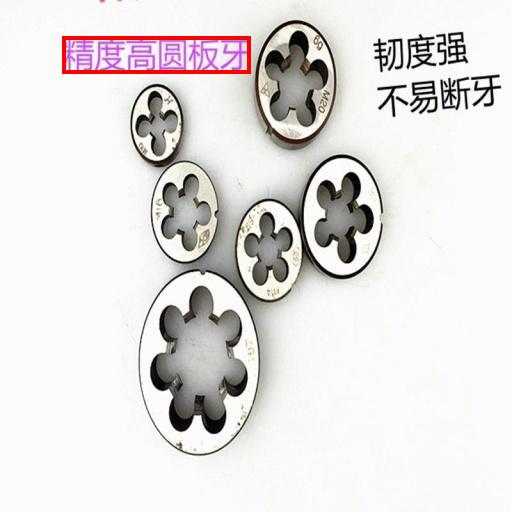}} &
\raisebox{-.5\height}{\includegraphics[width=0.16\textwidth]{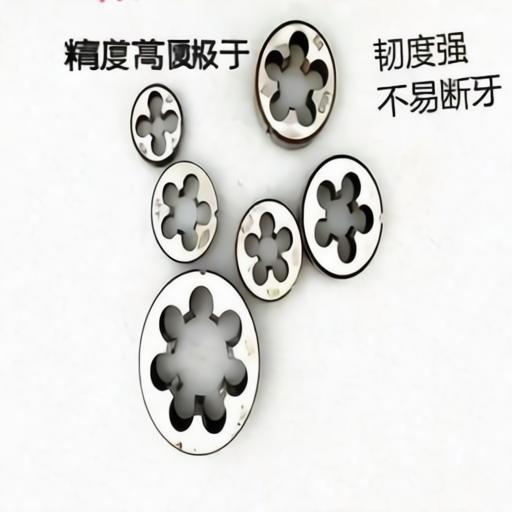}} &
\raisebox{-.5\height}{\includegraphics[width=0.16\textwidth]{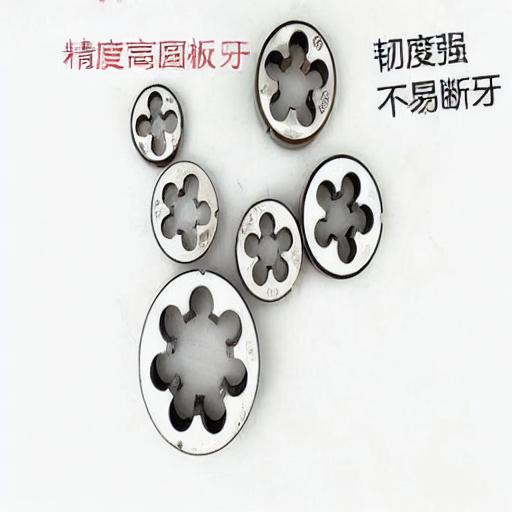}} &
\raisebox{-.5\height}{\includegraphics[width=0.16\textwidth]{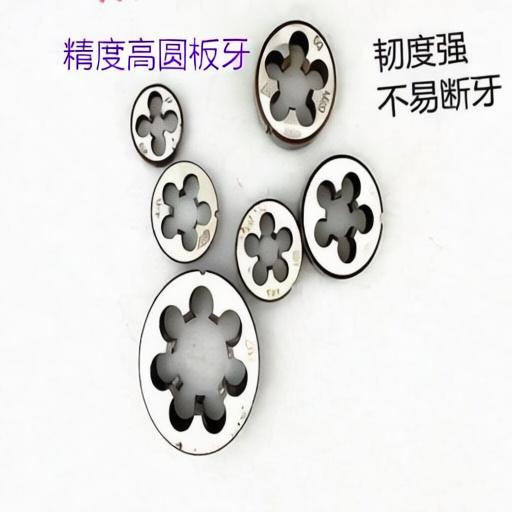}} \\
\raisebox{-.5\height}{
\shortstack{
\includegraphics{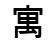} \\
\textit{(apartment)}}
} &
\raisebox{-.5\height}{\includegraphics[width=0.16\textwidth]{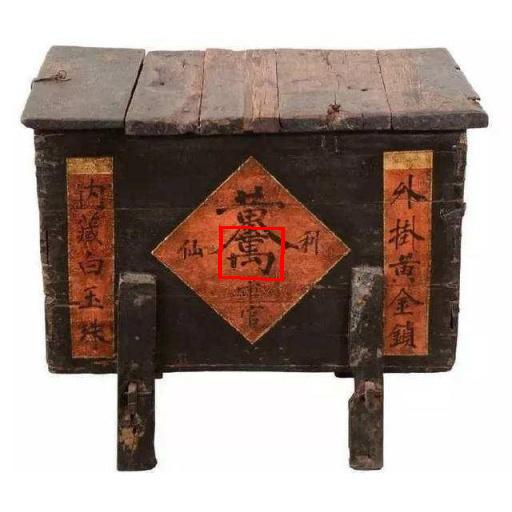}} &
\raisebox{-.5\height}{\includegraphics[width=0.16\textwidth]{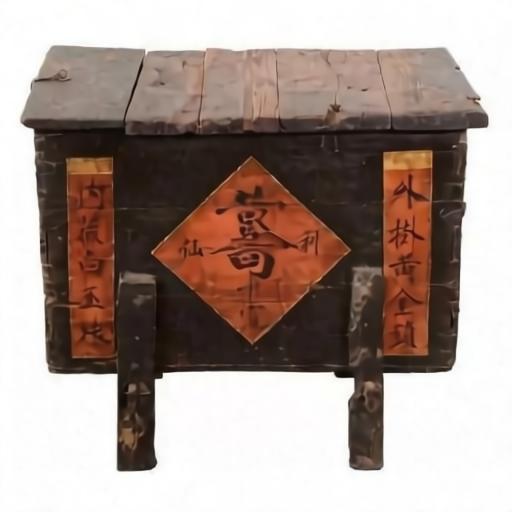}} &
\raisebox{-.5\height}{\includegraphics[width=0.16\textwidth]{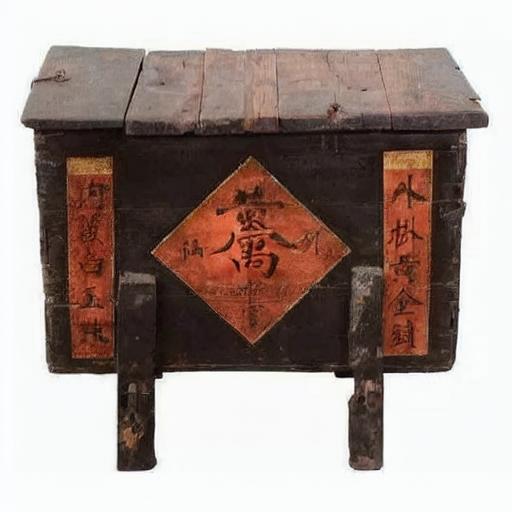}} &
\raisebox{-.5\height}{\includegraphics[width=0.16\textwidth]{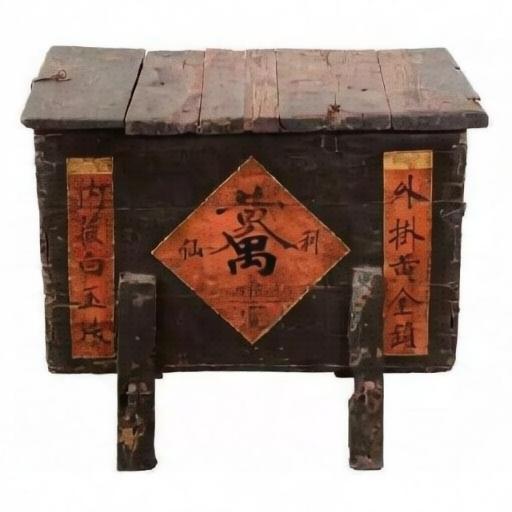}} \\
\end{tabular}
\caption{Comparison results on Chinese (Wukong) test set}
\label{fig:wukong}
\end{figure*}

\begin{figure*}[t]
\centering
\includegraphics[width=1\linewidth]{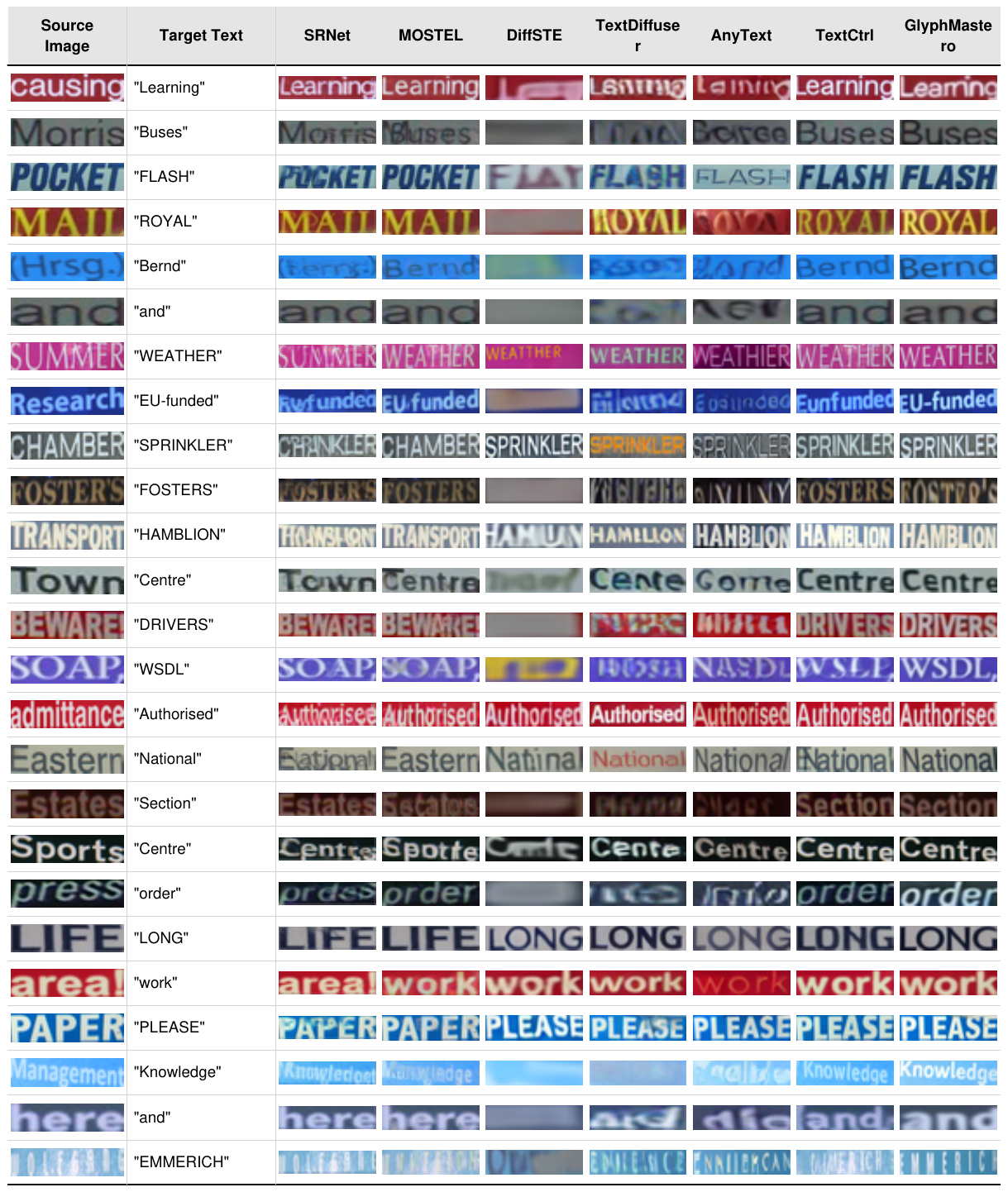}
\caption{Comparison of different scene text editing methods on the ScenePair dataset}
\label{scenepair_cmp}
\end{figure*}

We present additional qualitative comparisons in Figure \ref{fig:ours}, \ref{fig:laion}, \ref{fig:wukong}, and \ref{scenepair_cmp}. 

Figure~\ref{fig:ours} presents additional examples from our curated test set, demonstrating the efficacy of our approach for stylistic scene text editing. Additionally, Figures~\ref{fig:laion} and~\ref{fig:wukong} showcase random samples from the AnyText-Eval benchmark dataset - Figure~\ref{fig:laion} illustrates our model's performance on English text using the LAION dataset, while Figure~\ref{fig:wukong} highlights its capabilities with Chinese text from the Wukong dataset. Figure~\ref{scenepair_cmp} provides comparative results on TextCtrl's ScenePair test set as previously discussed. These qualitative results align consistently with our quantitative evaluations, validating the effectiveness of our proposed method.

\end{document}